\documentclass{article}

\usepackage{arxiv}
\usepackage{float}
\usepackage{multicol}
\usepackage{multirow}
\usepackage{esint}

\usepackage{amsmath} 
\usepackage{amssymb}
\usepackage{calc}

\usepackage[utf8]{inputenc} % allow utf-8 input
\usepackage[T1]{fontenc}    % use 8-bit T1 fonts
\usepackage{hyperref}       % hyperlinks
\usepackage{url}            % simple URL typesetting
\usepackage{booktabs}       % professional-quality tables
\usepackage{amsfonts}       % blackboard math symbols
\usepackage{nicefrac}       % compact symbols for 1/2, etc.
\usepackage{microtype}      % microtypography

\usepackage{graphicx}
\usepackage[
backend=biber,
style=nature,
]{biblatex}
\usepackage{doi}
\title{ShipGen: A Diffusion Model for Parametric Ship Hull Generation with Multiple Objectives and Constraints}

\author{ \href{https://orcid.org/0000-0001-9893-8619}{\includegraphics[scale=0.06]{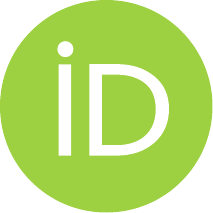}\hspace{1mm}Noah J.~Bagazinski}\thanks{Corresponding Author.} \\
	Department of Mechanical Engineering\\
	Massachusetts Institute of Technology\\
	Cambridge, MA 02139 \\
	\texttt{noahbagz@mit.edu} \\
	\And
	\href{https://orcid.org/0000-0002-5227-2628}{\includegraphics[scale=0.06]{orcid.pdf}\hspace{1mm}Faez ~Ahmed} \\
	Department of Mechanical Engineering\\
	Massachusetts Institute of Technology\\
	Cambridge, MA 02139 \\
	\texttt{faez@mit.edu}
}

\renewcommand{\shorttitle}{ShipGen}

\hypersetup{
pdftitle={ShipGen-DDPM},
pdfsubject={},
pdfauthor={Noah J.~Bagazinski, Faez~Ahmed},
pdfkeywords={Naval Architecture; Generative Artificial Intelligence; Deep Generative Models; Denoising Diffusion Probabilistic Model; DDPM;  Multi-objective Design; Design Constraint Satisfaction; Drag Reduction; Parametric Design; Ship Design}
}
\begin{document}

\maketitle

\begin{abstract}
Ship design is a years-long process that requires balancing complex design trade-offs to create a ship that is efficient and effective. Finding new ways to improve the ship design process can lead to significant cost savings in the time and effort required to design a ship and cost savings in the procurement and operation of a ship. One promising technology is generative artificial intelligence, which has been shown to reduce design cycle time and create novel, high-performing designs. In literature review, generative artificial intelligence has been shown to generate ship hulls; however, ship design is particularly difficult as the hull of a ship requires the consideration of many objectives. This paper presents a study on the generation of parametric ship hull designs using a parametric diffusion model that considers multiple objectives and constraints for the hulls. This denoising diffusion probabilistic model (DDPM) generates the tabular parametric design vectors of a ship hull, which is then constructed into a point cloud and mesh for performance evaluation. In addition to a tabular DDPM, this paper details adding guidance to improve the quality of generated parametric ship hull designs. By leveraging a classifier to guide sample generation, the DDPM produced feasible parametric ship hulls that maintain the coverage of the initial training dataset of ship hulls with a 99.5\% rate, a 149x improvement over random sampling of the design vector parameters across the design space. Parametric ship hulls produced with performance guidance saw an average of 91.4\% reduction in wave drag coefficients and an average of a 47.9x relative increase in the total displaced volume of the hulls compared to the mean performance of the hulls in the training dataset. The use of a DDPM to generate parametric ship hulls can reduce design time by generating high-performing hull designs for future analysis. These generated hulls have low drag and high volume, which can reduce the cost of operating a ship and increase its potential to generate revenue. 
\end{abstract}

% keywords can be removed

\section{Introduction}
Generative artificial intelligence (AI) models produce new instances of information that resemble the data used to train the model. While generative AI is famously known for generating text and image information, it can also be used to generate information to engineer products. Recent advances in generative AI provide promising new avenues to quickly generate designs. Including additional information in the training, such as a design's performance, can be leveraged to create designs with high performance. These advances are especially useful in the design of ships. Ship design currently requires a large team of naval architects to balance design trade-offs in a single ship's design. A generative AI model specifically trained to generate ship hulls can improve this workflow. Training such a model successfully is enabled by the availability of large datasets that include both design and performance information for ship hulls~\cite{bagazinski2023ship}. Hull design was chosen as a starting point for the generative model as the shape of the hull has a direct impact on over 70\% of the cost of a ship~\cite{lin2017feature}. It is also one of the first steps in the traditional workflow for ship design~\cite{evans1959basic}. The hull shape affects several key aspects of a ship's performance, including the buoyancy, upright stability, hydrodynamics, and general arrangements of the ship. With these considerations, the design of ship hulls provides an impactful avenue for the application of machine learning for engineering design. 

\begin{figure}[ht]
\begin{center}
\setlength{\unitlength}{0.012500in}%
\includegraphics[width = 5.5in]{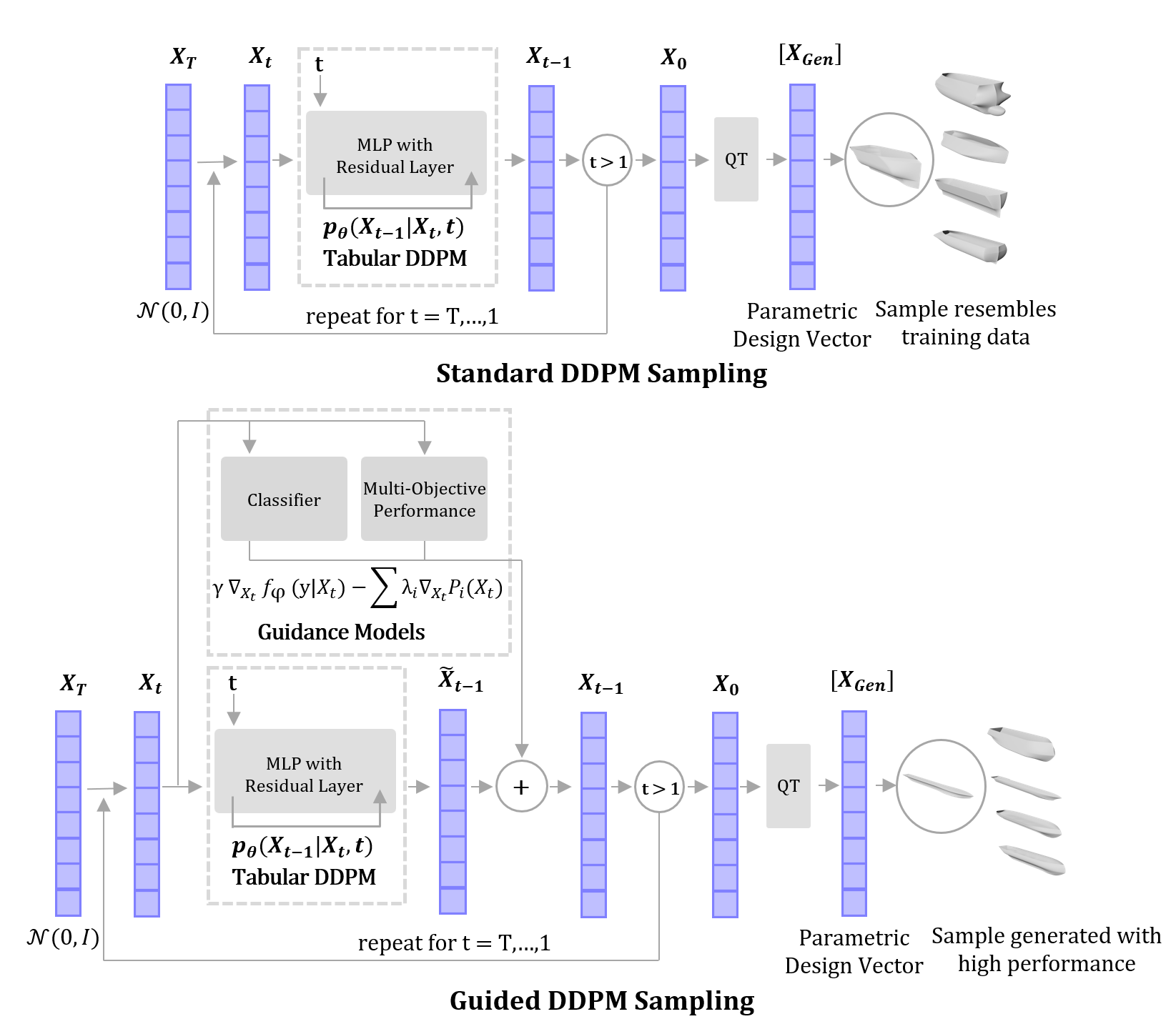}
\caption{Overview of utilizing a DDPM to generate parametric ship hull designs. When leveraging classifier and performance guidance from pre-trained neural networks, the DDPM is able to generate ship hull designs with high performance.}
\label{fig:figure_Overview} 
\end{center}
\end{figure}

A well-designed machine learning tool for ship design could learn design trade-offs for ships through the continual design and evaluation of many ship designs. This work demonstrates the use of a guided denoising diffusion probabilistic model (DDPM), a type of deep generative model, to rapidly generate high-performing and feasible parametric ship hull designs by generating parameters in a tabular format. This model, called ShipGen, generates early-stage hull designs considering seven performance metrics, creating shapes with low drag and high cargo-carrying capacity.  Figure~\ref{fig:figure_Overview} shows an overview of the work presented, highlighting that the implementation of classifier and performance guidance during the sampling process generates hulls with high performance. This work features model training with a publicly available dataset of parametric ship hulls, called ShipD~\cite{bagazinski2023ship}. The following sections detail the literature review of previous work, the methodology for creating and evaluating a tabular DDPM, the evaluation of ship hulls generated by the DDPM, and a discussion on the impact of the work. The hulls generated with the use of the guided DDPM are intended to be candidate designs for future analysis. As such, these generated hulls may not necessarily look exactly like realistic hull forms, but instead have design features that, in combination, lead to high performance. Through the development of the performance-guided DDPM for ship hull generation, the novel contributions of this paper are:
\begin{enumerate}
    \item The first known use of denoising 
diffusion probabilistic models for generating parametric tabular data for an engineering performance-focused design application.
    \item Showcase that classifier guidance in the DDPM navigates complex design feasibility constraints to generate feasible samples with over 99\% success while maintaining dataset design coverage. 
    \item Use of guidance to improve ship hull performance, with samples having an average 91.4\% reduction in wave drag coefficient and 47.9x more displaced volume compared to the mean performance of the hulls in the dataset.
\end{enumerate}

\section{Prior Work}
Generative AI for ship hull design was influenced by research in computational ship design and machine learning literature. The first subsection details prior work in computational ship design, including ship hull design representation, hull form design optimization, and the use of machine learning in ship design. The second subsection details the development of diffusion models and their applications in engineering design. 

\subsection{Computational Ship Design}
Computational Ship Design refers to the application of computer-based modeling, simulation, and optimization techniques in the design and analysis of marine vessels, facilitating more efficient, innovative, and integrated design solutions.
Historically, computational ship design can be divided into three categories: design representation, forward modeling which includes surrogate models, and inverse design or synthesis, which includes optimization methods. Recently, generative AI methods have emerged as a powerful technique, which could be used for the representation and synthesis of ship hull designs. 

In order to design a product with computational methods, the product needs to be represented in a way that a computer can understand. For ship design, the two most popular modes are parameterized vectors~\cite{brown2003multiple, HullOpt, read2009drag,  zhang2018parametric, chrismianto2014parametric, lu2016hydrodynamic, PSOShip_Opt,PSO_Multi_Opt, hodgesai, bagazinski2023ship}, and free form deformation techniques~\cite{wang2022shipEncoding,ao2021artificial,ao2022artificial,peri2001design,demo2021hull,abbas2023deepmorpher}. The benefit of using parameterized design representations for a hull is that the design is defined by a set of tunable parameters that both human designers and computers can interpret. 
The ease of use of parametric design representations has often limited the diversity of possible hull shapes. 
Conversely, FFD techniques present a different landscape. They allow for the creation of a broad array of shapes. Yet, these representations can be challenging for humans to interpret without a visual representation of the hull form. The works of Khan et al.~\cite{khan2022shape, khan2022geometric,khan2023shiphullgan},  Shaeffer et al.~\cite{shaeffer2023application, shaeffer2020application}, and Bagazinski et al.~\cite{bagazinski2023ship} have looked at various methods to create diverse design spaces and design datasets for ship hull design. These efforts aim to harness machine learning in ship hull design. 

In addition to design representation, computational design often has metrics for evaluating a design's performance. Finding computationally efficient methods for evaluating each generated design could lead to enhanced design generation. 
%In literature, one of the most commonly used performance metrics for ships is hydrodynamic drag. 
Hydrodynamic drag stands out as the predominant performance metric for ships in literature. Several rapid drag prediction techniques exist. Some, like Hollenbach's and Savitsky's methods, rely on statistical regressions from test data~\cite{hollenbach1998estimating,hollenbach2007efficient, savitsky1964hydro}. 
Other fast methods to predict wave drag are linear wave solvers, which provide accurate drag measurements with reduced computational effort relative to traditional computational fluid dynamics techniques. These solvers use potential flow to simulate the waves produced by a ship in a steady forward motion to estimate drag as a result of surface wave propagation. Different linear wave solvers include Michell's Integral~\cite{michell1898Wave,tuck1989wave}, Rankine Panel Methods~\cite{mantzaris1998rankine}, Neumann-Kelvin Theory (also called Dawson's Method)~\cite{dawson1977practical}, and Neumann-Michell Theory~\cite{noblesse2013neumann,yang2013practical, huang2013numerical}. These potential flow solvers input the 3D geometry of a hull and provide estimates of drag at typical operating speeds of a hull. The third method of creating a fast prediction of drag is to build a small dataset of drag measures to train a neural network to predict drag from a hull's design representation~\cite{ao2021artificial, ao2022artificial, khan2022geometric, khan2022shape, peri2001design, read2009drag, lu2016hydrodynamic, HullOpt,wang2022shipEncoding, marlantes2021Modeling, silva2023implementation}.

Combining hull design representation with efficient drag prediction equips designers with the tools needed for optimization algorithms. This enables hull design creation tailored for specific scenarios. A common objective in optimization literature is minimizing hull drag while adhering to geometric constraints. More recently, computational hull design has also been attempted using a tabular generative adversarial network (GAN) to quickly generate ship hull instances, that could be used for seeding populations for design optimization~\cite{khan2023shiphullgan}. The next improvement for generative AI in ship design is to implement a denoising diffusion probabilistic model (DDPM) for generating hull designs. Diffusion models provide improvements over GANs for generative design as DDPMs are more stable to train and provide superior sampling quality. Additionally, diffusion models can implement guidance without retraining the whole generative model. This way, new constraints or performance objectives can be integrated into design generation simply, whereas a GAN would need to be retrained for every new design consideration.~\cite{dhariwal2021diffusion}. For ship hull design, this means that a single model can be trained to generate high quality hulls that are tailored to specific user needs by integrating guidance models for different design considerations. This is particularly useful for ship design so that information from the design of many classes of ships can be considered in designing a ship hull. 

% Ideally, a proficient deep generative model can produce superior designs aligning with client requirements, bypassing the costly optimization with each new design set.
% The combination of hull design representation and fast drag prediction method provides the inputs needed to leverage an optimization algorithm to arrive at a hull design for a specific use case. Most of the literature reviewed for hull design optimization minimized the drag of a hull while maintaining some geometric constraints on the hull. One notable exception among the literature review for computational hull design is the development of a tabular generative adversarial network to quickly create instances of ship hulls. These generated hulls can initialize a population for design optimization~\cite{khan2023shiphullgan}. 
% In theory, a well trained deep generative model for design generation could produce high quality designs that meet customer needs without an expensive optimization process with every new set of design requirements. This work looks to expand on the possibility of high quality design generation without the expense of design optimization.

\subsection{Generative Design with Diffusion Models}
The transition from traditional design methods leads to a cutting-edge generative AI model: the denoising diffusion probabilistic model (DDPM). 
Gaining momentum in the machine learning domain, DDPMs iteratively modify a noisy data vector over many specified steps, transforming random data to mirror the statistics of training data~\cite{ho2020denoising}. The development of DDPMs in the last few years has shown that they are capable of generating complex data and already have applications for engineering design. For example, DDPMs were shown to create higher quality images as compared to generative adversarial networks~\cite{ho2020denoising}, a particularly difficult task as images are comprised of large patterns of pixels to visually represent something a human could see with their eyes. 

DDPMs work by training a neural network to predict small iterative denoising steps. The algorithm for training a diffusion model as defined by Ho et al. is in Table~\ref{table:DDPMTrainAlg}.
\begin{table} [H]
\begin{center}
\begin{tabular}{l l} 
\hline
1: & \textbf{repeat} \\
2: & $X_0 \sim q(X_0)$ \\
3: & $t \sim Uniform(\{1,...,T\})$ \\
4: & $\epsilon \sim N(0,I)$ \\
5: & Take gradient descent step on: \\
 & $\nabla_{\theta}||\epsilon - \epsilon_{\theta}(\sqrt{\bar{a_t}}X_0 + \sqrt{1 - \bar{a_t}}\epsilon, t)||^2$ \\
6: & \textbf{until} converged\\
\hline
\end{tabular}
\end{center}
\caption{This is the training algorithm for a standard DDPM. The DDPM is represented by the function $\epsilon_{\theta}(X_0,\epsilon,t)$ in step 5. }
\label{table:DDPMTrainAlg}
\end{table}

In the algorithm, the generated sample (design parameters) are represented by $X$, and noted with subscripts to indicate the denoising timestep. The DDPM itself is represented by $\epsilon_{\theta}$, indicating that the DDPM is trained to predict a small change in random noise across the vector.
Once trained, a DDPM generates samples by denoising a Gaussian noise vector over the predetermined timesteps. This results in samples that are within the training data's statistical distribution.  In the case of images, this could be a ``deep fake'' that looks like the training data. In the case of ship hull design, it could be a parameterized ship hull design. The sampling algorithm defined by Ho et al. is defined in Table~\ref{table:DDPMSampAlg}. 

\begin{table} [H]
\begin{center}
\begin{tabular}{l l} 
\hline
1: & $X_T \sim N(0,I)$\\
2: & \textbf{for} $t = T,...,1$ \textbf{do} \\
3: & $ Z \sim N(0,I)$ if $t > 1$, else $z = 0 $ \\
4: & $ X_{t-1} = \frac{1}{\sqrt{\alpha_t}} (X_t - \frac{1 - \alpha_t}{\sqrt{1 - \bar{\alpha_t}}} \epsilon_{\theta}(X_t ,t)) + \sigma_tZ $ \\
5: & \textbf{end for} \\
6: & \textbf{return} $X_0$ \\
\hline
\end{tabular}
\end{center}
\caption{This is the sampling algorithm for a standard DDPM. The DDPM is represented by the function $\epsilon_{\theta}(X_t,t)$ in step 4. }
\label{table:DDPMSampAlg}
\end{table}

% The next improvement to DDPMs was the implementation of guidance, where the gradients of a classifier neural network can be used to guide image synthesis towards satisfying a specific image classification label~\cite{dhariwal2021diffusion}. This led to the creation of text-to-image DDPMs that utilize guidance from text information to create custom and realistic images at will~\cite{ramesh2022hierarchical,rombach2021highresolution}. Guided DDPMs also have been implemented to generate 3D shapes from image data~\cite{liu2023zero1to3}. 
Subsequent advancements in DDPMs introduced guidance, where gradients from a classifier neural network guide image synthesis to match a specific image classification label~\cite{dhariwal2021diffusion}. This evolution birthed text-to-image DDPMs that employ text-based guidance to craft custom, lifelike images~\cite{ramesh2022hierarchical,rombach2021highresolution}. Guided DDPMs have found applications in generating 3D shapes from image data~\cite{liu2023zero1to3}.

Guided diffusion can be applied to engineering design generation. For example, guided diffusion has been used to create two-dimensional structures~\cite{maze2022topodiff,giannone2023aligning,giannone2023learning} and vehicles~\cite{arechiga2023drag} using image data. In these instances, the guidance of the design generation by image-based DDPMs is applied to constraint satisfaction and improved performance. DDPMs can generate high-quality designs, navigate complex constraints, and implement precise generation with guidance, which makes them an excellent deep generative model for designing ship hulls. The subsequent sections demonstrate a tabular DDPM to generate parametric ship hull designs that give improved performance through the implementation of guidance.

%%%%%%%%%%%%%%%%%%%%%%%%%%%%%%%%%%%%%%%%%%
\section{Methods}
This section outlines the methodology behind developing a guided DDPM for ship hull design. This section explores the ship hull dataset, delves into tabular DDPMs, and introduces both classifier and performance guidance for sampling ship hulls with a DDPM. A secondary methods section on conditional DDPMs is included in the Appendix.

\subsection{Ship-D Dataset and Hull Parameterization}
The Ship-D dataset consists of 30,000 parameterized ship hulls. The hulls are parameterized with 45 terms. These terms are applied to a set of algebraic equations to define and characterize the surface of the hull. These terms were construed through analyzing and characterizing the shape and curvature of many different publicly available hull geometries. 
The parameters cover various aspects:
\begin{itemize}
    \item Principal dimensions (e.g., overall length, beam at main deck)
    \item Cross-section of the parallel midbody (e.g., deadrise angle, chine radius)
    \item Geometry of bow and stern taper
    \item Geometry of bulbs at bow and stern
\end{itemize}

\begin{figure}[H]
\begin{center}
\setlength{\unitlength}{0.012500in}%
\includegraphics[width = 5.5in]{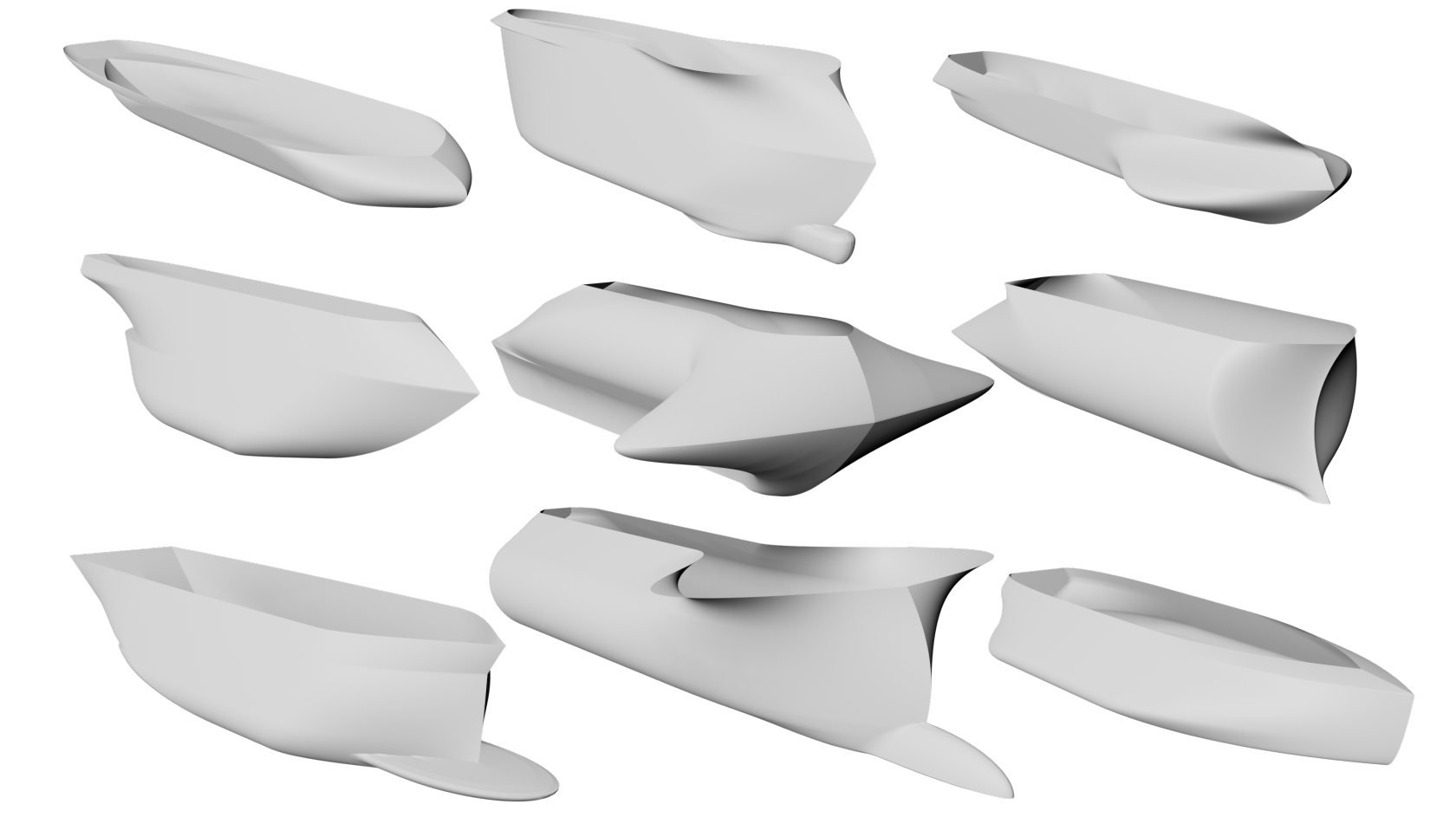}
\caption{A selection of hulls from the Ship-D dataset, showing the variability possible with the hull parameterization. A random sampling from the dataset may lead to unrealistic hulls, containing combinations of features that do not resemble real-world ships and features that lead to poor performance.}
\label{fig:figure_Ship-DHulls} 
\end{center}
\end{figure}

These parameters, designed to capture a range of curvature and shapes, encompass the features seen in a diverse variety of vessels from large ships to smaller boats. Their dual design facilitates human understanding and computer-generated input. Full documentation of the hull design parameters is provided at ~\url{https://decode.mit.edu/projects/ShipGen/}. Additionally, Figure~\ref{fig:figure_PARAMETERS} in the Appendix lists the parameters and provides details so human designers can create parametric hulls with this representation. A glimpse into the Ship-D dataset is provided in Figure~\ref{fig:figure_Ship-DHulls}, showcasing the diverse shapes achievable with the parametric design scheme. As these designs are randomly sampled across the entire feasible design space, they do not necessarily look like realistic hull designs. The performance of these hulls was not considered in their random sampling. Many of these hulls are relatively low performing: having high drag, low displacement volumes, and high surface area. The feasibility criteria used to generate these hulls are described in the next section.

\subsubsection{Feasibility Constraints for Hull Geometry}
%Copy and pasted from Ship D paper
While the parameterization can define a large design space of hull geometries, constraints on the parameterization are needed to ensure that a feasible hull will be produced by a specific set of parameters. To satisfy a ``feasible'' hull shape, the hull's surface only needs to satisfy two criteria: 
\begin{enumerate}
    \item The hull is watertight, meaning that there are no holes on its surface.
    \item The hull surface is not self-intersecting.     
\end{enumerate}
As the hull surface is defined by a set of equations with constants dictated by the parameter values, conditions to determine whether a hull's surface satisfies the two main feasibility criteria can be solved algebraically. The advantage to algebraically solving these conditions is significantly reduced computational effort to check hull feasibility with the algebraic constraints compared to feasibility checks with mesh generation. After searching through the design space of the hull parameterization and examining the equations that define the hull surface, a set of forty nine constraints were defined to determine if a hull surface produced from a specific parameterization satisfies the two feasibility criteria. Figure~\ref{fig:figure_CONSTRAINTS} in the Appendix lists the 49 algebraic constraints and provides information on each of their satisfaction conditions.

Conversely, the two feasibility criteria can be checked by constructing the mesh of a hull and analyzing its surface. Mesh generation and feasibility checks are computed in $O(Nlog(N))$, where $N$ is the number of vertices on the mesh. On an Intel Core i9-10980XE processor, the construction and check of a hull mesh with approximately 80,000 vertices is 1.77 seconds. Comparatively, the algebraic constraints check the design feasibility of a parametric hull in 0.000199 seconds. This is a ten-thousand-fold increase in speed for checking design feasibility with the algebraic constraints. A uniform random sampling of the design parameters leads to generation of a feasible hull in approximately 1 per 150 tries. In addition to the 30,000 feasible hulls in the Ship-D dataset, an additional 20,000 design vectors (called invalid samples) that violate at least one feasibility constraint were generated. These invalid samples are used to train models in classifying and distinguishing between feasible and infeasible design vectors~\cite{maze2022topodiff}.

\subsubsection{Hull Performance Measures}
The Ship-D dataset already contains ten geometric measures and thirty-two wave drag calculations for each hull. The ten geometric measures allow naval architects to characterize a hull when designing a ship. The ten geometric measures are calculated using trapezoidal integration at ten draft marks spaced along the depth of the hull. These ten geometric measures are:
\begin{enumerate}
    \item Height of draftmark
    \item Length of the waterline
    \item Area of the waterplane
    \item Surface area of the hull below the specified draftmark (wetted surface)
    \item Longitudinal centers of flotation (waterplane centroid)
    \item Second moment of area about the longitudinal axis of the waterplane
    \item Second moment of area about the transverse axis of the waterplane
    \item Displaced volume below the draftmark 
    \item Longitudinal center of buoyancy
    \item Vertical center of buoyancy
\end{enumerate}
As these metrics have the units of length to some power $L^n$, they are normalized by the first term in the parameterization, $LOA$, to its respective power. For example, lengths are normalized by $LOA$, areas by $LOA^2$, volumes by $LOA^3$, and area moments of inertia by $LOA^4$. This allows computational analysis on the geometry of the hull to be performed independently of the hull's scale. 

In addition to the geometric measures, the Ship-D dataset has thirty-two wave drag coefficients for each hull across four different drafts and eight velocity conditions. The four drafts are 25\%, 33\%, 50\%, and 67\% of the hull's total depth. The eight velocity conditions are normalized using Froude scaling. The eight velocities are between $F_n =0.10$ and $F_n = 0.45$ in increments of 0.05, corresponding to typical operating conditions of traditional displacement hulls~\cite{AppNavArch,newman2018marine}. The Froude number is the relative scaling between inertial and gravitational forces described in the equation below:
\begin{equation}
F_n = \frac{U}{\sqrt{gL}}
\label{eq_Fn}
\end{equation}
 Where $U$ is the hull speed, $g$ is gravity and $L$ is a length scale. The length used in simulating the 32 speed-draft conditions of the hulls was the length of the waterline at the tested draft mark. This way, thirty two unique conditions were measured. Wave drag is both a function of the hulls geometry, and the hydrodynamics of waves propagating off of the hull from it's forward motion. Including a full spectrum of speed-draft conditions in the dataset allows a machine learning model to learn the effects of drag due to changing submerged geometry with draft and speed. This provides significantly more information relating to the geometry and performance of a hull than available by measuring a single operating condition. This allows a generative model using the Ship-D dataset to generalize wave drag in the design process. As the generative model is intended to produce conceptual hull designs, it is imperative that the exact speed-draft condition be unknown so that the model generates hulls that generally have low drag. The future work section will detail goals for generating hull designs tailored to specific use cases, which could include specific speed-draft conditions. 
 
 The Michell Integral was chosen to simulate wave drag over other linear wave methods for its relative computational efficiency and the accuracy it provides. The Michell integral is a linear estimate of the wave drag of a slender ship in forward motion. It is defined by the following equation~\cite{michell1898Wave,tuck1989wave}: 

% \begin{equation}
%  R_w = \frac{A \rho g^2}{\pi U^2} \int_{1}^{\infty} (I^2 + J^2)* \frac{\lambda^2}{\sqrt{\lambda^2 - 1}} d\lambda
% \label{eq_Mich}
% \end{equation}
\begin{equation}
 R_w = \frac{A \rho g^2}{\pi U^2} \int_{1}^{\infty} (I^2 + J^2) \frac{\lambda^2}{\sqrt{\lambda^2 - 1}} \, d\lambda
\label{eq_Mich}
\end{equation}

 where $\rho$ is the density of water, $g$ is gravitational acceleration, $U$ is the ship's speed, and $A, I, and J$ are integrated terms relating to the surface normal across the hull and the direction of wave propagation. Further insight into these terms is in Michell's paper form 1898~\cite{michell1898Wave}.
 
 In addition to scaling the relative speed and draft conditions for the hulls, the wave drag is also scaled to ensure consistency across the dataset:
%  \begin{equation}
%   C_w = \frac{R_w}{\frac{1}{2} \rho U^2 LOA^2}
% \label{eq_Cw}
% \end{equation}   
\begin{equation}
  C_w = \frac{R_w}{\frac{1}{2} \rho U^2 \cdot \text{LOA}^2}
\label{eq_Cw}
\end{equation}

Typical drag coefficients of hulls are scaled by the wetted surface area of the hull. Within the dataset, however, the wetted surface area of the hulls can vary greatly. Instead, the Length-Overall (LOA) is used instead as this is the first term in the parameterization. For the purposes of applying machine learning using the dataset, the wave drag coefficient can be characterized by the remaining 44 terms in the parameterization and the hull's relative speed and draft. 

An additional two measures of the hulls are included in this paper and will be added to the Ship-D dataset. The first measure is the Gaussian curvature of the hull's surface. The second metric is a measure of the largest rectangular prism that can be vertically lowered into the hull, referred to as MaxBox for the remainder of this paper. Gaussian curvature quantifies the double curvature of a surface. The average Gaussian curvature is calculated for these hulls to assess the manufacturing complexity of the hull's surface. As most large ships are constructed from welded sheet steel or aluminum, bending a sheet along two principal axes of curvature is a difficult task for both the sheet forming process and for welding the edge of a complex surface to another. By measuring the average double curvature of each hull, an understanding of the difficulty of manufacturing the hull surface is gained for the dataset. The Gaussian curvature is calculated for the hulls using a finite difference method to measure the principal curvature of the hull in the YZ plane and in the XY plane for a uniform grid of points on the hull~\cite{dalle2006comparison}. Equation~\ref{eq_GC} calculates the average Gaussian curvature over the surface of the hull. The terms $R_{XY}$ and $R_{YZ}$ are the radii calculated using the finite difference method along the two principal axes of hull's surface.  Gaussian curvature has units $1/L^2$ and is hence normalized by $LOA^2$

% \begin{equation}
%  GC = \frac {\oiint_{S} \frac{dA} {R_{XY}(x,y,z) * R_{YZ}(x,y,z)}} {Total Surface Area}
% \label{eq_GC}
% \end{equation}

\begin{equation}
 GC = \frac {\oiint_{S} \frac{\mathrm{d}A} {R_{XY}(x,y,z) \cdot R_{YZ}(x,y,z)}} {\text{Total Surface Area}}
\label{eq_GC}
\end{equation}

The MaxBox measures the box with maximum volume that is completely inscribed by the hull that can be vertically lowered into the hull through the waterplane at the hull's top deck. This provides a measure for evaluating a candidate region within the ship hull for allocating cargo holds. Additionally, as the MaxBox is open at the deck of the ship, a crane can service this entire volume within each hull. The MaxBox for each hull was optimized with a Nelder-Mead simplex optimization algorithm to maximize the volume of the box while constrained by the surface of the hull and the waterplane of the top deck~\cite{nelder1965simplex}. Included in the dataset is the forward (X) position of the box, its length, width, depth, and volume. These results are normalized by their length dimensionality, $1/LOA$ and $1/LOA^3$.

Among the available performance measures in the Ship-D dataset, seven were selected to be implemented in the performance-driven design generation of ship hulls. The goal of selecting the seven performance metrics that generally describe the quality of a hull. These metrics provide an avenue to compare hulls directly to each other with useful characteristics that consider the hulls' hydrodynamics, hydrostatics, and manufacturability. These seven performance metrics are: 

\begin{enumerate}
    \item Aggregated sum of wave drag coefficients.
    \item Surface area of the hull up to 50\% of its total depth
    \item Total surface area of the hull 
    \item Displaced volume of the hull up to 50\% of its total depth
    \item Total displaced volume of the hull
    \item Volume of the MaxBox
    \item Gaussian curvature
\end{enumerate}
The aggregated sum of wave drag coefficients was selected as a way to quickly characterize the general wave drag of a given hull. In large ships, wave drag is the primary component in the ship's total drag. By learning how a ship's hull shape affects drag, a generative AI model could generate hulls with low wave drag, saving ship operation costs through reduced fuel consumption. The aggregated sum of the wave drag coefficients is defined in Equation~\ref{eq_NormCW}. It is important to note that this performance metric and five of the other metrics are all represented on a logarithmic scale. Due to the geometry of the hull designs these performance metrics span several orders of magnitude across the Ship-D dataset. The distribution of the logarithmic scaled performance metrics is normal, a desired quality for machine learning.
% \begin{equation}
%  C_{w*} = \sum_{i=1}^{32} Log_{10}(C_w{}_i)
% \label{eq_NormCW}
% \end{equation}
\begin{equation}
 C_{w*} = \sum_{i=1}^{32} \log_{10}(C_{w_i})
\label{eq_NormCW}
\end{equation}

The surface area of the lower half of the hull was selected as a performance measure as this is the portion of a hull's surface that is most likely to be submerged when the hull is in water. The wetted surface of the ship affects the viscous drag acting on the hull. Reducing the wetted surface area of a hull can reduce the total drag of a ship, saving operation costs through reduced fuel consumption. Additionally, the total surface area of the hull was selected as this can consider the amount of material needed to manufacture the surface of the ship hull. By reducing the total surface area of the ship,  manufacturing costs can be reduced by generating hull designs with less total surface area. The two measures of surface area are provided in Equation~\ref{eq_NormSA50} and Equation~\ref{eq_NormSA100}.

% \begin{equation}
%  SA_{50\%*} = Log_{10} (\frac{\int_{0}^{T/D_d = 0.5} \delta SA(z) \delta z} {LOA^2})
% \label{eq_NormSA50}
% \end{equation}

% \begin{equation}
%  SA_{100\%*} = Log_{10} (\frac{\int_{0}^{T/D_d = 1.0} \delta SA(z) \delta z} {LOA^2})
% \label{eq_NormSA100}
% \end{equation}

\begin{equation}
 SA_{50\%*} = \log_{10} \left( \frac{\int_{0}^{T/D_d = 0.5} \delta SA(z) \, \delta z} {LOA^2} \right)
\label{eq_NormSA50}
\end{equation}

\begin{equation}
 SA_{100\%*} = \log_{10} \left( \frac{\int_{0}^{T/D_d = 1.0} \delta SA(z) \, \delta z} {LOA^2} \right)
\label{eq_NormSA100}
\end{equation}

The displaced volume of the bottom half of the hull was selected as it characterizes the portion of the hull that contains much of the displaced volume for buoyant forces. This performance metric therefore characterizes the relative total weight of the ship and its cargo. Learning how the hull design parameters affect the displaced volume of the hull can lead to generating hull designs that can carry more weight. The total displaced volume of the hull was also selected as this measure characterizes the total volume capacity available for cargo, outfitting, and other systems on the ship. Learning how the ship hull design parameters affect its total displaced volume, ship hulls with greater total volume can be generated. These two measures of volume affect the ability of ships to generate revenue through the shipment of cargo. The two volume measures are calculated with Equation~\ref{eq_NormV50} and Equation~\ref{eq_NormV100}. With the intention of maximizing the volume metrics in hull generation, the volume measures are multiplied by -1. This conforms the volume maximization problem to a ``minimization'' problem akin to the other performance objectives. 

% \begin{equation}
%  V_{50\%*} = -Log_{10} (\frac{\int_{0}^{T/D_d = 0.5} \delta V(z) \delta z} {LOA^3})
% \label{eq_NormV50}
% \end{equation}

% \begin{equation}
%  V_{100\%*} = -Log_{10} (\frac{\int_{0}^{T/D_d = 1.0} \delta V(z) \delta z} {LOA^3})
% \label{eq_NormV100}
% \end{equation}

\begin{equation}
 V_{50\%*} = -\log_{10} \left( \frac{\int_{0}^{T/D_d = 0.5} \delta V(z) \, \delta z} {\text{LOA}^3} \right)
\label{eq_NormV50}
\end{equation}

\begin{equation}
 V_{100\%*} = -\log_{10} \left( \frac{\int_{0}^{T/D_d = 1.0} \delta V(z) \, \delta z} {\text{LOA}^3} \right)
\label{eq_NormV100}
\end{equation}

An additional measure of volume is the MaxBox volume. As described earlier in this section, the MaxBox metric measures the ratio of the most useful cargo-carrying volume of the ship compared to the hull's total displaced volume. Learning how the design parameters affect MaxBox can lead to the generation of hulls with more useful cargo-carrying capacity. This can also lead to greater revenue through a ship's operation. MaxBox is not on a logarithmic scale like the other measures and it is calculated with Equation~\ref{eq_NormMaxBox}.

% \begin{equation}
% MaxBox_* = -\frac{Volume_{MaxBox}} {Volume_{T/D_d = 1.0}}
% \label{eq_NormMaxBox}
% \end{equation}
\begin{equation}
\text{MaxBox}_* = -\frac{\text{Volume}_{\text{MaxBox}}} {\text{Volume}_{T/D_d = 1.0}}
\label{eq_NormMaxBox}
\end{equation}

The final performance metric selected is the average Gaussian curvature of the hull. Since Gaussian curvature is a measure of a hull's surface complexity, it affects the manufacturing costs of a ship. Reducing the average Gaussian curvature of a ship hull can lead to reduced manufacturing costs, making it a critical metric for a ship hull. The average Gaussian curvature is normalized for machine learning using Equation~\ref{eq_NormGC}.

% \begin{equation}
%  GC_* = Log_{10}(GC * LOA^2)
% \label{eq_NormGC}
% \end{equation}
\begin{equation}
 GC_* = \log_{10}(GC \cdot LOA^2)
\label{eq_NormGC}
\end{equation}

While the aforementioned seven metrics were chosen to demonstrate the efficacy of the proposed methodology, it is crucial to highlight that this is not an exhaustive list of performance measures for ship hull evaluation. Indeed, a significant strength of the proposed diffusion model lies in its adaptability. It allows users to integrate additional performance metrics without necessitating retraining. This flexibility underscores the model's robustness and its potential to be tailored to various specific needs, optimizing designs based on a myriad of performance criteria.

\subsection{Dataset Coverage and Generated Sample Evaluation}
In order to characterize the DDPMs' abilities to cover the total parametric dataset space and generate feasible designs, two measures are utilized throughout the remainder of the paper. To visually characterize how a set of generated hull designs covers the dataset space of the Ship-D hulls, a two-dimensional principal component analysis is trained with the Ship-D parametric design vectors. When evaluating the designs generated with DDPMs, the PCA of the generated samples is plotted against the PCA of a random selection of the Ship-D dataset hulls. This shows the relative spread of the generated designs compared to the dataset hulls. In addition to visualization, coverage and realism quantify a model's ability to generate samples similar to the training dataset.
\begin{itemize}
    \item Coverage is quantified as the mean chamfer distance of each dataset instance from its nearest neighbor among the generated samples.
    \item Realism, on the other hand, measures the mean chamfer distance of each generated sample instance from its closest match within the dataset~\cite{regenwetter2023beyond}.
\end{itemize}

 Chamfer distance is the Euclidean distance between two hull design vectors.  For two sets of parameterized hull designs, $A$ and $B$, the Chamfer distance finds the distances from a design vector in $A$ to its nearest neighbor in $B$. The distance metric used is the squared Euclidean distance between the two vectors. The formula for this evaluation metric is:

% \begin{equation}
% CD = ||A_n - B_{n*}||^2 
% \label{eq_CD}
% \end{equation}
\begin{equation}
CD = \Vert A_n - B_{n*} \Vert^2 
\label{eq_CD}
\end{equation}

where $B_{n*}$ is the nearest neighbor of the design vector $A_n$ in $B$. Chamfer distance is normalized for coverage and realism with the following equation:
%\begin{equation}
%CD_* = \frac {\frac{1}{N_A} * \sum_{n=1}^{N_A}(CD_n) - CD_{worst-case}} {CD_{best-case} - CD_{worst-case}}
%\label{eq_CD_norm}
%\end{equation}
\begin{equation}
CD_* = \frac{\frac{1}{N_A} \sum_{n=1}^{N_A} CD_n - CD_{\text{worst-case}}}{CD_{\text{best-case}} - CD_{\text{worst-case}}}
\label{eq_CD_norm}
\end{equation}

where $N_A$ is the number of points in set $A$ and $CD_*$ is a normalized Chamfer distance, being coverage or realism. As a baseline, coverage is linearly normalized between the best- and worst-case scenarios derived from an analysis of the dataset. The best-case coverage is the mean nearest neighbor distance of each hull in the dataset (100\% coverage), which is equal to 4.315. The worst-case coverage is the mean chamfer distance between the centroid of the dataset hulls and each hull in the dataset, (0\% coverage), which is equal to 26.930. realism is also linearly normalized to be between the minimum and maximum chamfer distance between the dataset hulls and their nearest neighbor.

To benchmark the DDPMs' ability to generate feasible samples, two studies on the feasibility of hulls generated by interpolating between hull design vectors were conducted. The first study generated sample hulls by finding the midpoint between two random hull design vectors from the Ship-D dataset. The second study generated sample hulls by finding the midpoint between a random hull and its nearest neighbor in the Ship-D dataset. The results of these studies are provided in the Results Section.

An additional benchmark study was conducted using a tabular generative adversarial network called CTGAN~\cite{ctgan}. The CTGAN was trained to generate feasible hull designs, only gathering information from the 30,000 feasible hull designs in the Ship-D dataset. The goal of this benchmark study is to compare the ability of the CTGAN and the DDPM to generate feasible hulls and cover the dataset without explicitly identifying feasible or infeasible hull designs for the model. In training, the CTGAN learns the parametric information encoded in the design vectors and generates samples to match the distribution of the dataset samples. The results of this study are included in Table~\ref{table:gammaCoverage} and Table~\ref{table:gammaFease}.

When leveraging classifier guidance with a DDPM to generate feasible samples, the dataset coverage is greatly affected by the same hyperparameter that influences the feasibility of generated samples. 
% Two metrics, coverage and realism are explored in this paper. coverage is calculated as the inverse of the distance each instance of the dataset is from its nearest neighbor belonging to the generated samples. realism is the opposite of this: the inverse of the distance each instance of the generated samples are from its nearest neighbor belonging to the dataset members~\cite{regenwetter2023beyond}.

\subsection{Denoising Diffusion Probabilistic Models}
A denoising diffusion probabilistic model (DDPM) is a generative AI model that generates new instances of data by denoising random information over many steps, so that the generated sample falls within the statistical distribution of the training dataset samples. A tabular DDPM was built and trained on the ship hull parametric design information from the Ship-D dataset. The DDPM used to create ShipGen was inspired by the work of Kotelnikov et.al, called TabDDPM~\cite{kotelnikov2023tabddpm}. Unlike popular image-focused diffusion models, this DDPM is trained on tabular information to generate tabular information.  Prior to training, the parametric design vectors were transformed with a quantile normalizer to re-scale the distribution of the design parameters to have a normal distribution with the same mean and variance as the parameters in the dataset. A second linear transformation re-scaled the bounds so that the range of each parameter exists between -1 and 1.  These transformations ensure that the parametric design data is fit for the tabular DDPM. Training this model provided a baseline to verify that the tabular DDPM produces ship hulls with parametric information within the relative distribution of the Ship-D dataset. The Results Section provides the results of both the parameter distribution and feasibility constraint satisfaction of these generated samples. 

\subsubsection{Standard Diffusion Model}
The standard DDPM follows the training and sampling algorithms defined in Table~\ref{table:DDPMTrainAlg} and Table~\ref{table:DDPMSampAlg}. The standard DDPM implicitly learns the statistical relationships between the parameters in each sample. In the sampling process, the trained DDPM generates samples that are statistically similar to the designs in the dataset. There is no extra consideration for design feasibility or design performance. Since the Ship-D dataset is comprised of randomly sampled hulls that meet the feasibility criteria, any increase in the DDPMs ability to produce feasible hulls compared to pure random sampling is due to the DDPM implicitly learning the relationships between the design parameters that lead to feasible hull designs. 

\begin{figure}[ht]
\begin{center}
\setlength{\unitlength}{0.012500in}%
\includegraphics[width = 5.5in]{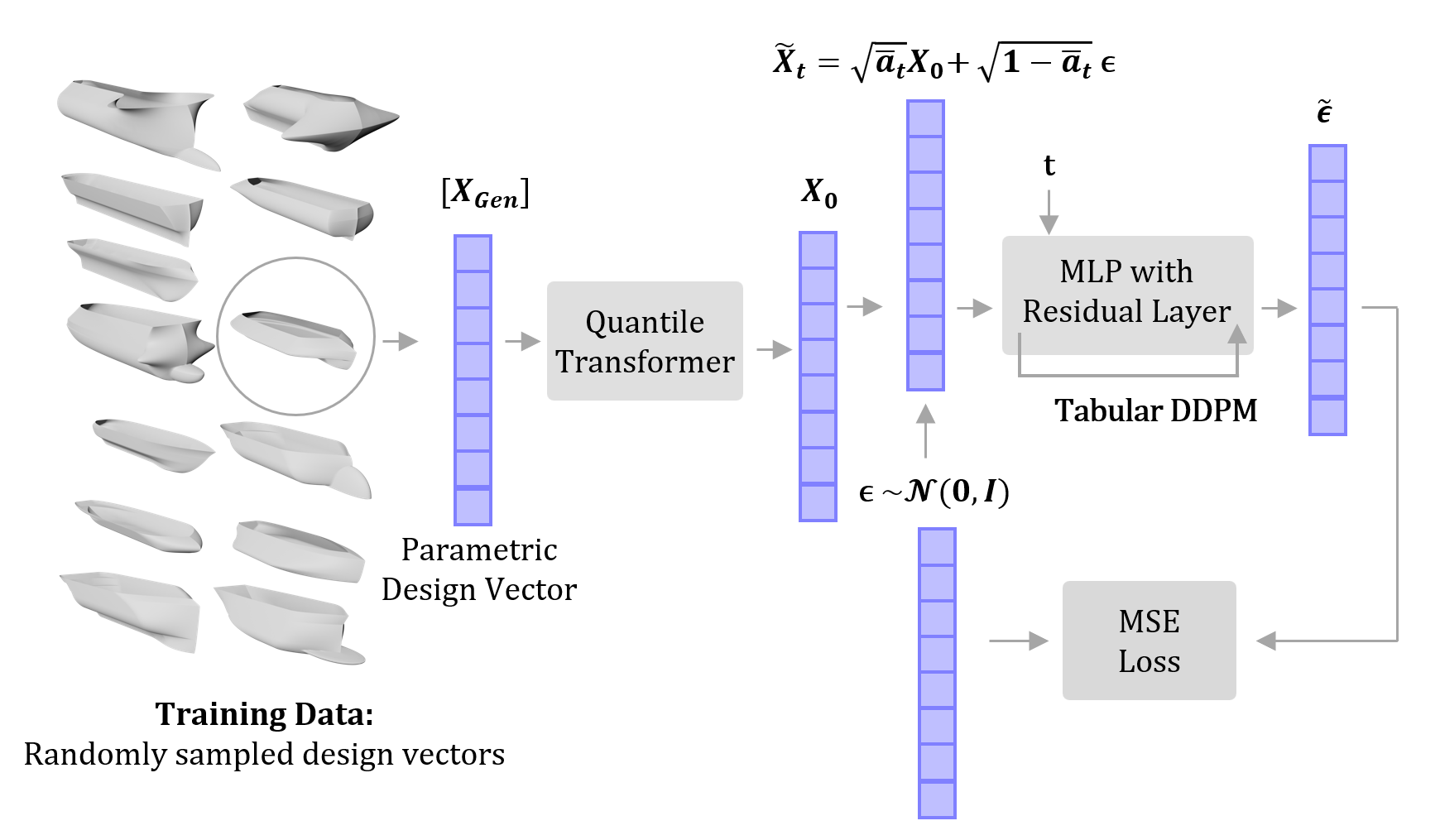}
\caption{During training, the DDPM predicts a denoising step, given a timestep embedding and a partially noised sample design vector.}
\label{fig:figure_DDPM_Training} 
\end{center}
\end{figure}

During training, a feasible design vector is quantile normalized and partially noised according to the training algorithm in Table~\ref{table:DDPMTrainAlg}. Then the DDPM predicts a noised vector given the timestep embedding and the partially noised vector. The mean squared difference between the predicted noise vector and a pure noise vector is the loss of this prediction. The mean squared loss then back-propagates through the DDPM to update its weights and biases. This process is repeated for the 30,000 feasible design vectors across one thousand denoising timesteps in random batches to train the DDPM. Figure~\ref{fig:figure_DDPM_Training} illustrates the training process for one design vector at one timestep.

\begin{figure}[ht]
\begin{center}
\setlength{\unitlength}{0.012500in}%
\includegraphics[width = 5.5in]{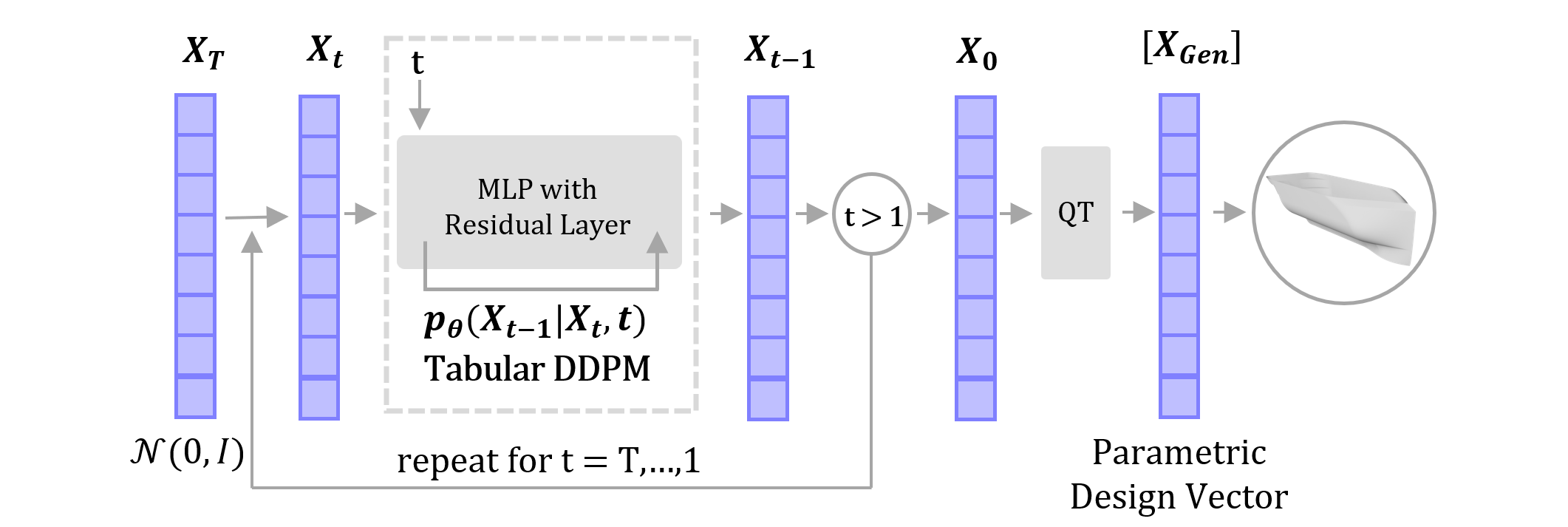}
\caption{During sampling, the standard DDPM denoises a vector over one thousand timesteps, generating a sample design vector that statistically aligns with the training data.}
\label{fig:figure_DDPM_Sampling} 
\end{center}
\end{figure}

After training, the standard DDPM can sample new design vectors. The initial seed for sampling is a Gaussian noise vector of size N, where N is the number of design parameters. The DDPM denoises this vector one thousand times, taking into account the timestep embedding at each iteration. After the denoising process, the final denoised vector is reverse-quantile normalized so that it becomes a design vector fitting the Ship-D parametric design scheme. This generated design vector can then be checked for feasibility constraint satisfaction. If the design is feasible, a point cloud and mesh of the hull is generated to evaluate the sample's performance. Figure~\ref{fig:figure_DDPM_Sampling} illustrates the sampling process for a single design vector. 

\subsubsection{Classifier Guidance for Diffusion Models}
An additional method of influencing sample generation towards feasibility constraint satisfaction is with classifier guidance~\cite{dhariwal2021diffusion}. Classifier guidance leverages the gradients of a trained design classifier during the standard DDPM's sampling process to influence a design to meet a certain classifier label. In this case, the classifier label characterizes whether a design is feasible or infeasible. Here, the 30,000 Ship-D design vectors and the 20,000 infeasible design vectors trained a classifier to predict design feasibility. At each timestep in the sampling process, the gradient of the trained classifier for a target class, $f_{\phi}(y|X_t)$, with respect to the parameterized design vector, $X_t$, is calculated. This gradient is multiplied by a hyperparameter, $\gamma$, and is added to the sample during Step 4 of the DDPM sampling algorithm defined in Table~\ref{table:DDPMSampAlg}. A classifier guided DDPM is created by replacing Step 4 with Equation~\ref{eq_ClassGuideSample}. 
% \begin{equation}
%  X_{t-1} = \frac{1}{\sqrt{\alpha_t}} (X_t - \frac{1 - \alpha_t}{\sqrt{1 - \bar{\alpha_t}}} \epsilon_{\theta}(X_t ,t)) + \sigma_t(Z(1-\gamma)) + \gamma \nabla_{X_t} f_{\phi}(y|X_t) 
% \label{eq_ClassGuideSample}
% \end{equation}
\begin{equation}
 X_{t-1} = \frac{1}{\sqrt{\alpha_t}} \left(X_t - \frac{1 - \alpha_t}{\sqrt{1 - \bar{\alpha}_t}} \epsilon_{\theta}(X_t ,t)\right) + \sigma_t(Z(1-\gamma)) + \gamma \nabla_{X_t} f_{\phi}(y|X_t) 
\label{eq_ClassGuideSample}
\end{equation}

The Results Section provides data from tuning $\gamma$, as it will be shown that the hyperparameter has an effect on both the likelihood of producing feasible hull design vectors and on the distribution of these vectors relative to the Ship-D dataset. Without the need to additionally train the standard DDPM itself, adding guidance gradients in the sampling process can be accomplished easily. Figure~\ref{fig:figure_GuidedSampling} and Figure~\ref{fig:figure_GuidedDiffusion} illustrate classifier guidance used in conjunction with performance guidance to generate parameterized hull designs with high performance. The next subsection details the addition of more guidance models to generate hulls while considering the hull's performance. 

\subsubsection{Performance Guidance for Diffusion Models}
Similar to classifier guidance, a neural network trained to predict the performance of a hull can also be used to guide sample generation. Seven residual neural networks were trained to predict the normalized performance of a hull given its parametric design vector. The 30,000 feasible hull designs in the ShipD dataset were used for the training data. The performance prediction neural networks all have the same structure: 4 hidden layers with 256 nodes, where the first hidden layer is added as a  residual to the final hidden layer. The normalization of the performance metrics distributes them over a Gaussian, improving the prediction accuracy of the neural network. 

During sampling, the gradients of the normalized performance prediction from the neural networks is used to guide the DDPM's sampling process. The performance gradient of each of the objectives, 
$\nabla_{X_t}P_i(X_t)$ is multiplied by a weight, 
$\lambda_i$. While generating samples, the weights of the $\lambda$ values for each performance objective are normalized so that they are positive and sum to 1.0 for each sample. This way, a broad spectrum of samples is generated with unique combinations of weighted influences from the seven performance objectives. Performance guidance is achieved by replacing Step 4 in the DDPM sampling algorithm with Equation~\ref{eq_PerfGuideSample}.
% \begin{equation}
%  X_{t-1} = \frac{1}{\sqrt{\alpha_t}} (X_t - \frac{1 - \alpha_t}{\sqrt{1 - \bar{\alpha_t}}} \epsilon_{\theta}(X_t ,t)) + \sigma_t(Z(1-\gamma)) + \gamma \nabla_{X_t} f_{\phi}(y|X_t)  - \sum_{i=1}^{7}{\lambda_i\nabla_{X_t}P_i(X_t)}
% \label{eq_PerfGuideSample}
% \end{equation}
\begin{equation}
 X_{t-1} = \frac{1}{\sqrt{\alpha_t}} \left(X_t - \frac{1 - \alpha_t}{\sqrt{1 - \bar{\alpha}_t}} \epsilon_{\theta}(X_t ,t)\right) + \sigma_t(Z(1-\gamma)) + \gamma \nabla_{X_t} f_{\phi}(y|X_t)  - \sum_{i=1}^{7}{\lambda_i\nabla_{X_t}P_i(X_t)}
\label{eq_PerfGuideSample}
\end{equation}

\begin{figure}[ht]
\begin{center}
\setlength{\unitlength}{0.012500in}%
\includegraphics[width=5.5in]{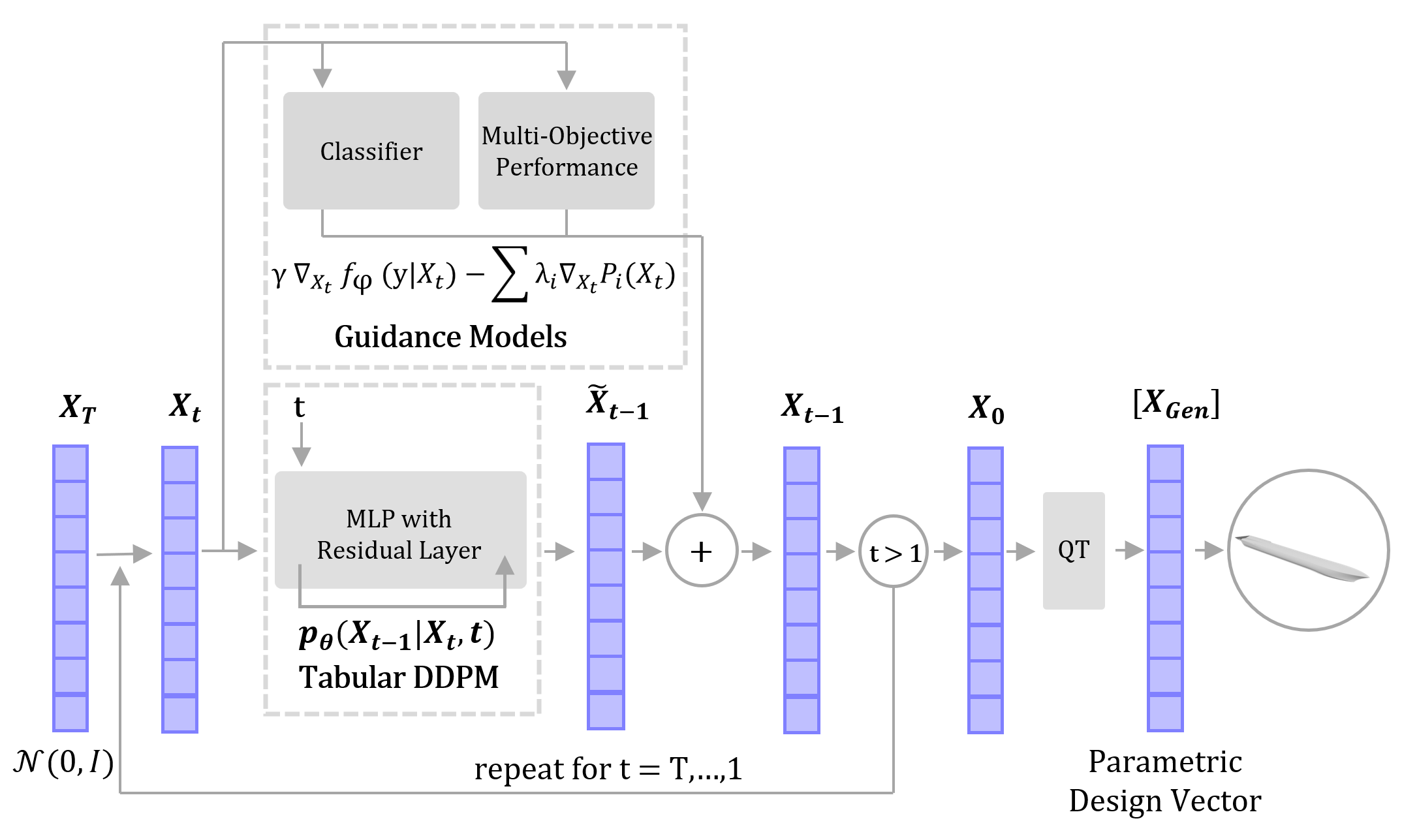}
\caption{For guided sampling, generated vectors are denoised with the standard DDPM at each timestep and then influenced with guidance gradients in each denoisng iteration.}
\label{fig:figure_GuidedSampling} 
\end{center}
\end{figure}

During sampling, the gradients of both the classifier and performance prediction models are calculated at each timestep. Weighting these gradients with $\gamma$ and $\lambda$ influences the impact each individual model has on the sampling process. The classifier guidance weight, $\gamma$, is set equal to 0.5 so that both a high degree of sample diversity and sample feasibility are maintained. Figure~\ref{fig:figure_GuidedSampling} shows how guidance from the classifier and performance prediction models are implemented into the denoising process. 

\begin{figure}[ht]
\begin{center}
\setlength{\unitlength}{0.012500in}%
\includegraphics[width=5.5in]{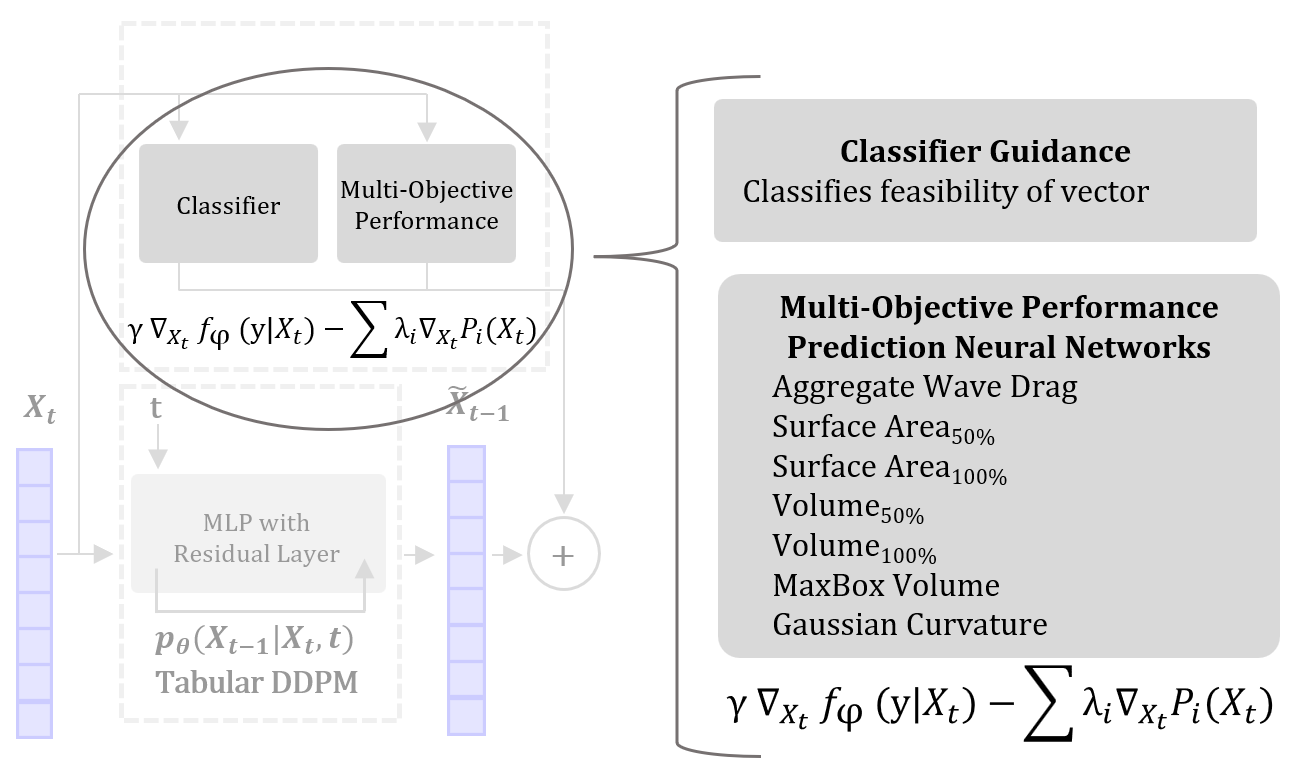}
\caption{Classifier and performance guidance is the result of leveraging gradients of pre-trained models to influence the denoising process of the DDPM. The figure highlights the models implemented for this experiment.}
\label{fig:figure_GuidedDiffusion} 
\end{center}
\end{figure}

Figure~\ref{fig:figure_GuidedDiffusion} highlights the contributions of both the feasibility classifier and the performance prediction neural networks in guidance. For performance guidance, the gradients of the seven performance prediction networks are calculated for each sampling timestep for $X_t$. Then, the gradients are weighted by their respective $\lambda$ value. The sum of the weighted gradients is subtracted from the output of the standard DDPM to create the next partially denoised vector, $X_{t-1}$, in the sampling process. The gradients are subtracted to follow the scheme of ``minimizing'' the performance objectives in generated samples.   

After sampling, the generated design vectors are checked for feasibility. The performance of the feasible generated hulls are then calculated with the same simulations used to create the original dataset. In the Results Section, Table~\ref{table:PerfPredictionInfo} will showcase the mean normalized performance between the dataset hulls and the generated hulls, as well as a scale factor between the true performance of the two sets of hulls. The Results Section also provides data on the fit of these performance regression networks.

%%%%%%%%%%%%%%%%%%%%%%%%%%%%%%%%%%%%%%%%%%
\section{Results}
This section contains the results of the studies described in the Methods Section. The first subsection provides results on the feasibility and design spread of parameterized hulls generated with interpolation between existing hulls in the Ship-D dataset. The second subsection gives the results of generating feasible hulls using a standard tabular DDPM and with a guided DDPM. The third subsection provides the results on generating hulls using performance guidance, including the results of the performance prediction residual neural networks. The Appendix contains the results of training a conditional DDPM with both feasible and invalid hulls.

\subsection{Benchmark Feasibility Constraint Satisfaction Studies}
An initial study generating hulls using interpolation methods was conducted to measure the success rate of generating feasible hulls using the Ship-D dataset. The first study generated thirty thousand parametric hulls by interpolating the parameters halfway between two random hulls belonging to the Ship-D dataset. This interpolation method generated feasible hulls at a rate of 93.1\%, listed in Table~\ref{table:gammaFease}. The second interpolation method generated hulls by interpolating between a dataset hull and its nearest neighbor hull. The second interpolation method generated feasible hulls, with a success rate of 93.8\%. Table~\ref{table:gammaCoverage} lists the dataset coverage of these two interpolation methods. The first interpolation method maintains a normalized coverage ratio of 0.965 compared to the baseline coverage, while the second interpolation method exceeds the baseline coverage, having a ratio of 1.059. These two interpolation methods will serve as benchmarks for the feasibility and dataset coverage analysis on hulls generated with the DDPM.

The CTGAN benchmark study provided a baseline for a trained generative model to generate feasible hulls implicitly. The CTGAN was trained on the 30,000 feasible hull designs Ship-D dataset to implicitly learn the combinations of parameter values that define ``feasibility''. This study generated thirty thousand samples of hulls and measured the dataset coverage and feasibility of these samples. The sample coverage was decreased with the CTGAN, having a normalized coverage ratio of 0.94 compared to the baseline. This is slightly reduced from the interpolation benchmark coverage ratios of 0.965 and 1.059. The coverage measures for the CTGAN study are included in Table~\ref{table:gammaCoverage}. Of the CTGAN generated samples, only 0.7\% satisfy all feasibility constraints. This finding is included in Table~\ref{table:gammaFease}. The CTGAN is only marginally better than randomly sampling the design space to create a feasible parametric hull design. Further analysis of the CTGAN benchmark study is included in the Discussion Section.

\subsection{Feasibility Constraint Satisfaction with Tabular Denoising Diffusion Probabilistic Models}
This subsection provides the design feasibility and dataset coverage of samples generated with different types of DDPMs. The types of DDPMs considered for sample generation are the standard DDPM and guided DDPM. The following subsections provide the results for each type of DDPM.

\subsubsection{Standard DDPM Leads to Good Feasibility and Coverage}
\begin{figure}[H]
\begin{center}
\setlength{\unitlength}{0.012500in}%
\includegraphics[width=3.5in]{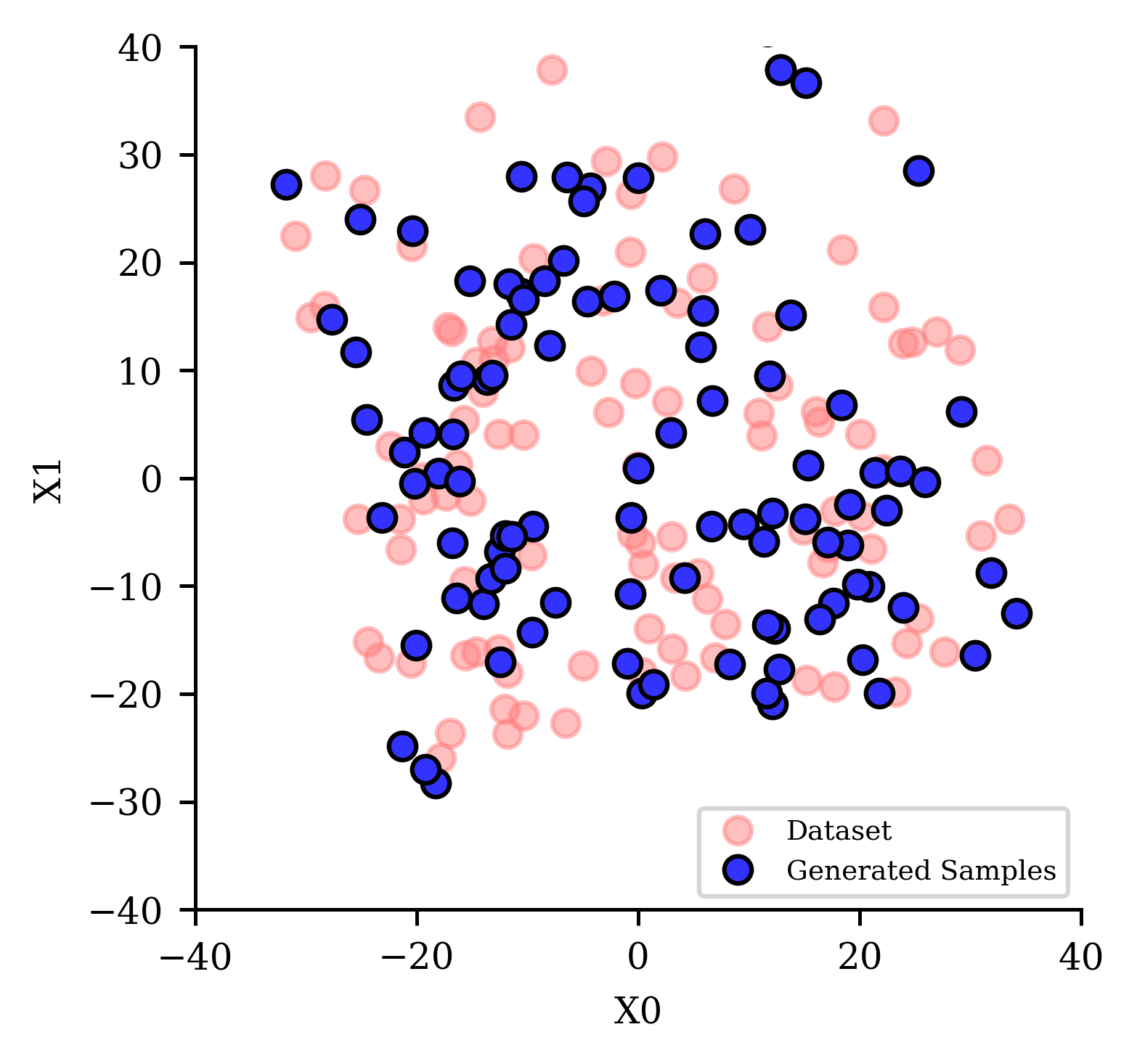}
\caption{Two-dimensional Principal Component Analysis of the hull parameterization shows that hulls generated with a standard DDPM maintain most of the dataset coverage.}
\label{fig:figure_diff_vanilla_PCA} 
\end{center}
\end{figure}

A standard DDPM is trained only on the parametric design information from the dataset. Samples generated from a standard DDPM are made up of the implicit statistical relationships learned from the parameters in feasible hulls. The standard DDPM produces feasible hulls 51.1\% of the time, as seen in Table~\ref{table:gammaFease}. Throughout the Results Section, a two-dimensional principal component analysis (PCA) is used to illustrate the spread of generated sample hulls compared to the Ship-D dataset hulls. The PCA is trained on the parametric hull design data from the Ship-D dataset and is used to transform generated samples into the two-dimensional PCA for visualization. Figure~\ref{fig:figure_diff_vanilla_PCA} shows that the standard DDPM generates samples that maintain most of the dataset coverage, maintaining a normalized coverage ratio of 0.984. The dataset coverage and feasibility of samples created with the standard diffusion model are included in Table~\ref{table:gammaCoverage} and Table~\ref{table:gammaFease}.

%\begin{figure}[H]
%\begin{center}
%\setlength{\unitlength}{0.012500in}%
%\includesvg{images/VanillaDiffusionConstraintSatisfaction.svg}
%\caption{Bar graph showing the number of individuals generated with an increasing number of constraint violations. Leveraging a standard DDPM to generate %hull parameterization leads to hull parameterizations that satisfy all constraints 49.5\% of the time}
%\label{fig:figure_diff_vanilla_bar} 
%\end{center}
%\end{figure}

\subsubsection{Guided Denoising Diffusion Probabilistic Model for Enhanced Feasibility}
\begin{figure}[ht]
\begin{center}
\setlength{\unitlength}{0.012500in}%
\includegraphics[width=3.5in]{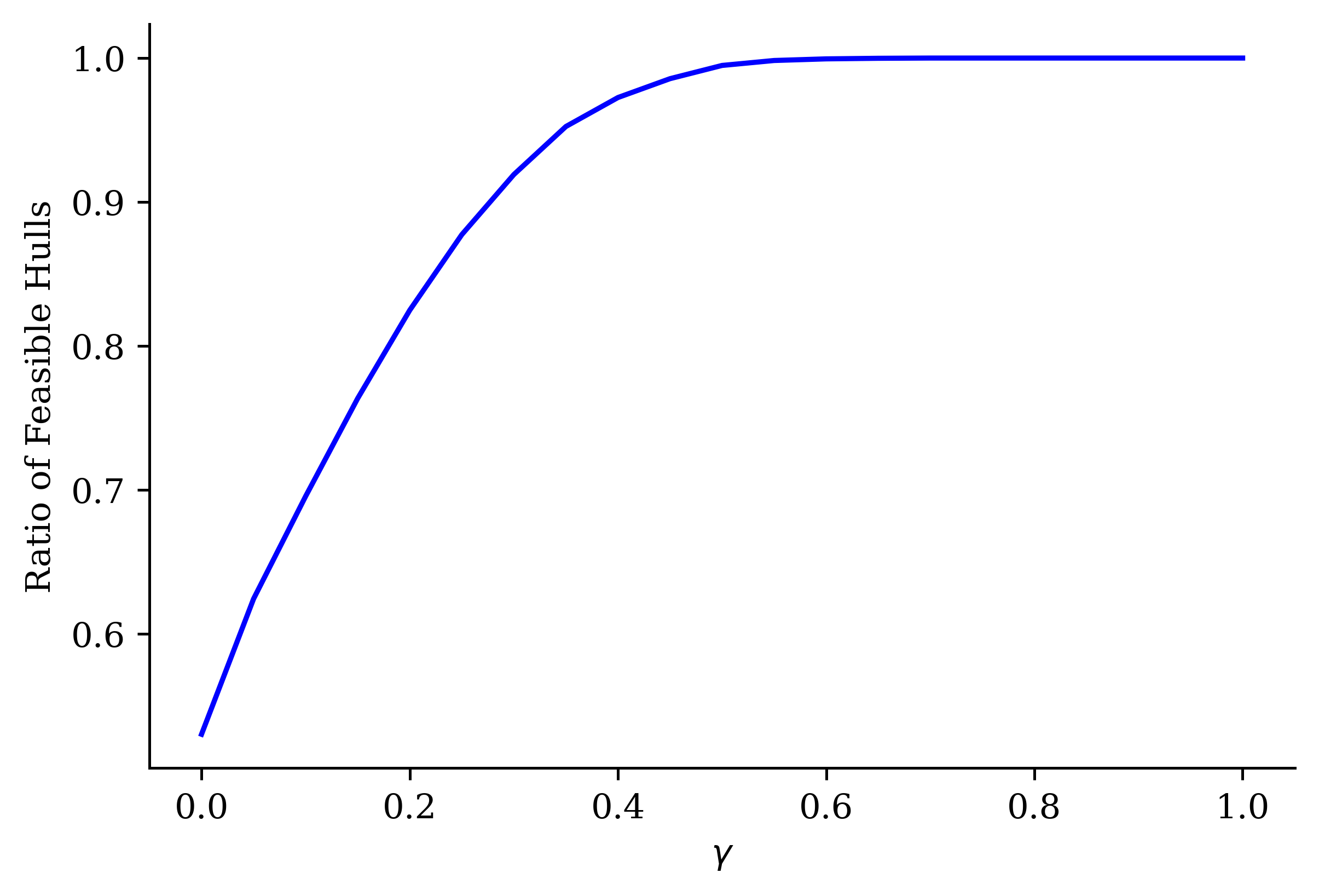}
\caption{Hull design feasibility is highly dependent on $\gamma$ in guided diffusion. The percentage of feasible generated hulls is above 90\% when $\gamma$ is greater than 0.3}
\label{fig:figure_feasVsGamma} 
\end{center}
\end{figure}

\begin{figure}[ht]
\begin{center}
\setlength{\unitlength}{0.012500in}%
\includegraphics[width=3.5in]{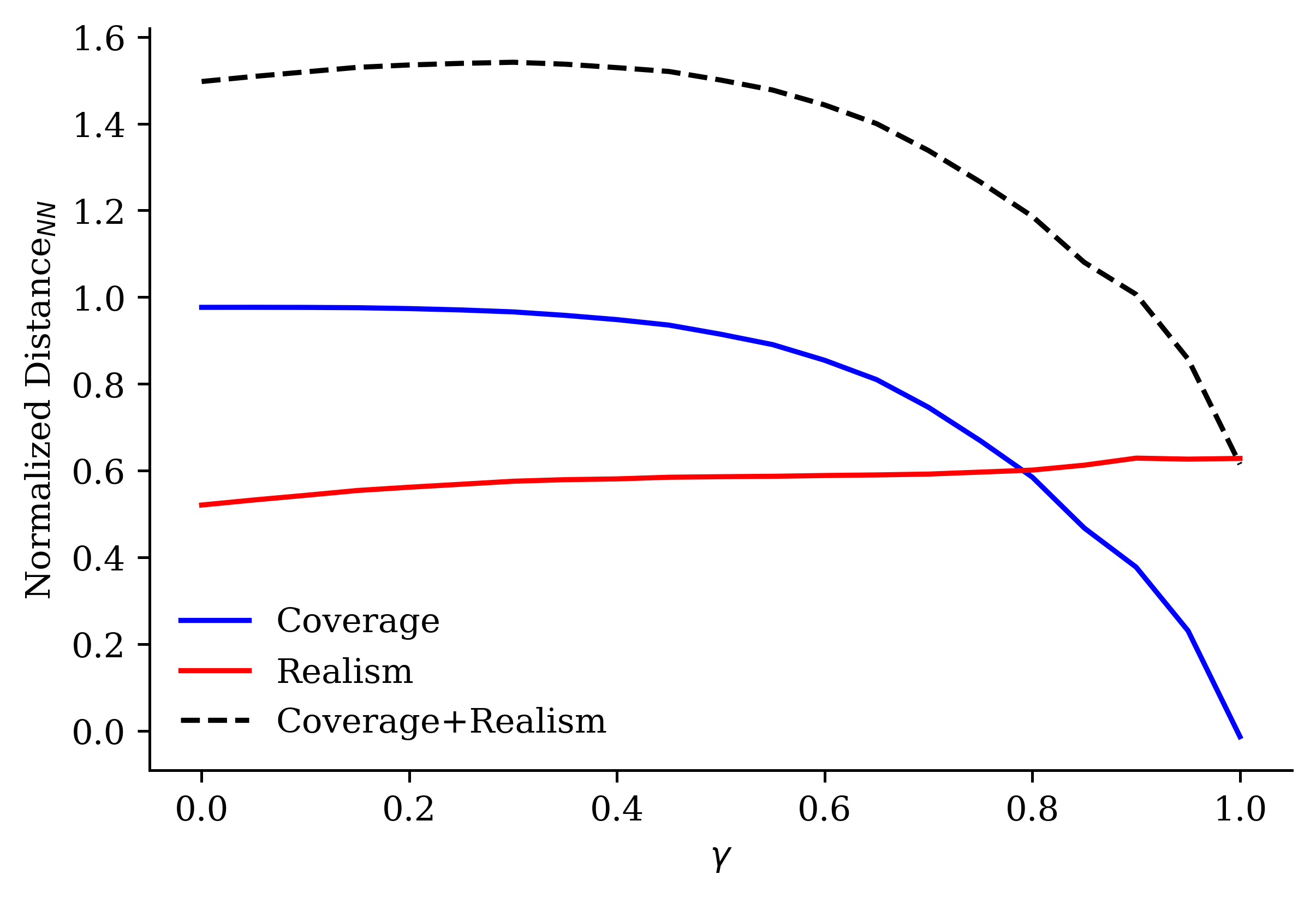}
\caption{Realism and coverage of the generated samples is strongly affected by $\gamma$. When $\gamma$ is approximately 0.5, the sum of realism and coverage is maximized.}
\label{fig:figure_CandRVsGamma} 
\end{center}
\end{figure}

\begin{figure}[ht]
\begin{center}
\setlength{\unitlength}{0.012500in}%
\includegraphics[width=5.5in]{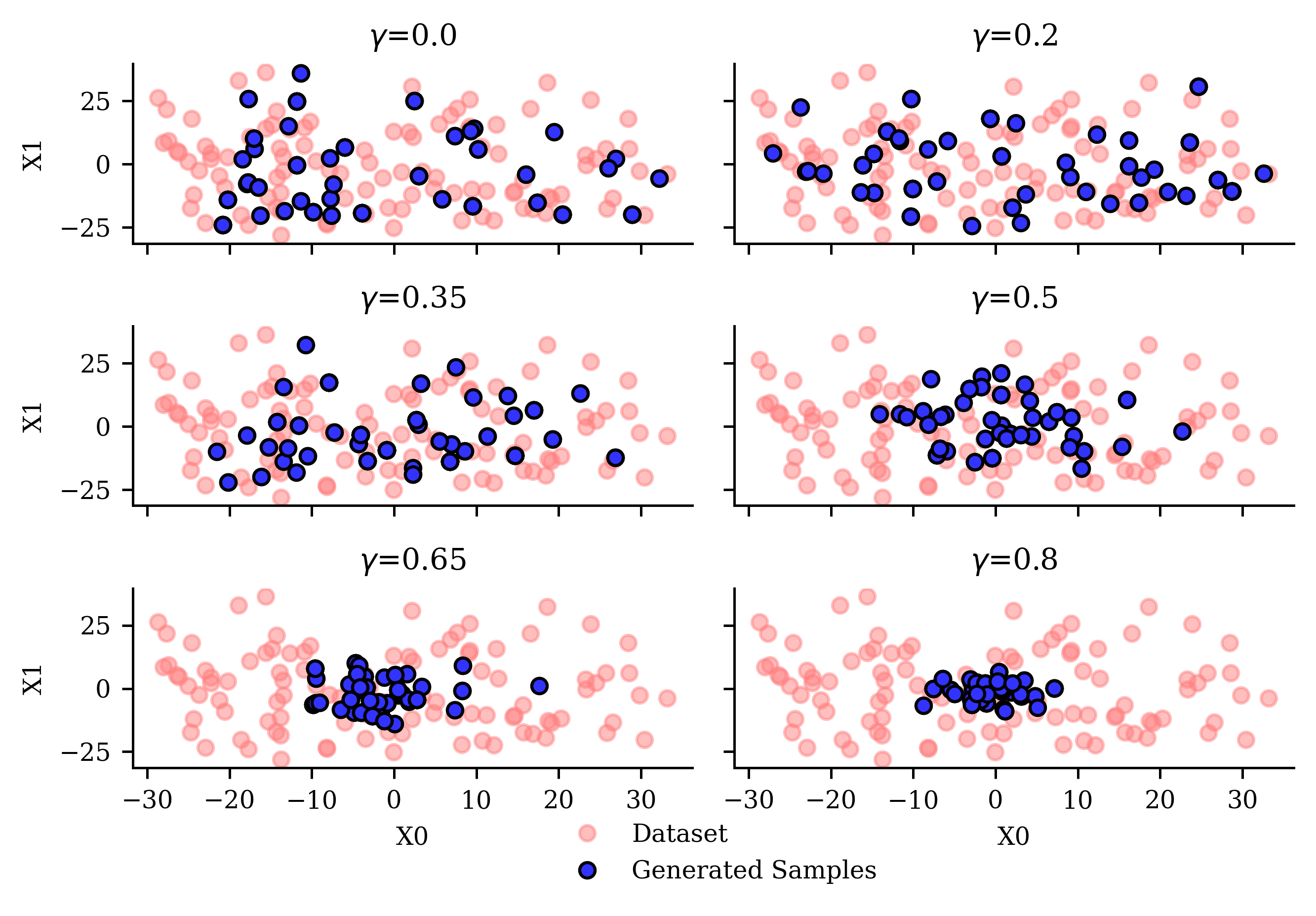}
\caption{Two-dimensional Principal Component Analysis of the hull parameterization shows that  the dataset coverage is reduced by increasing the hyperparameter, $\gamma$}
\label{fig:figure_diff_guided_gamma_PCA} 
\end{center}
\end{figure}

\begin{figure}[ht]
\begin{center}
\setlength{\unitlength}{0.012500in}%
\includegraphics[width=3.5in]{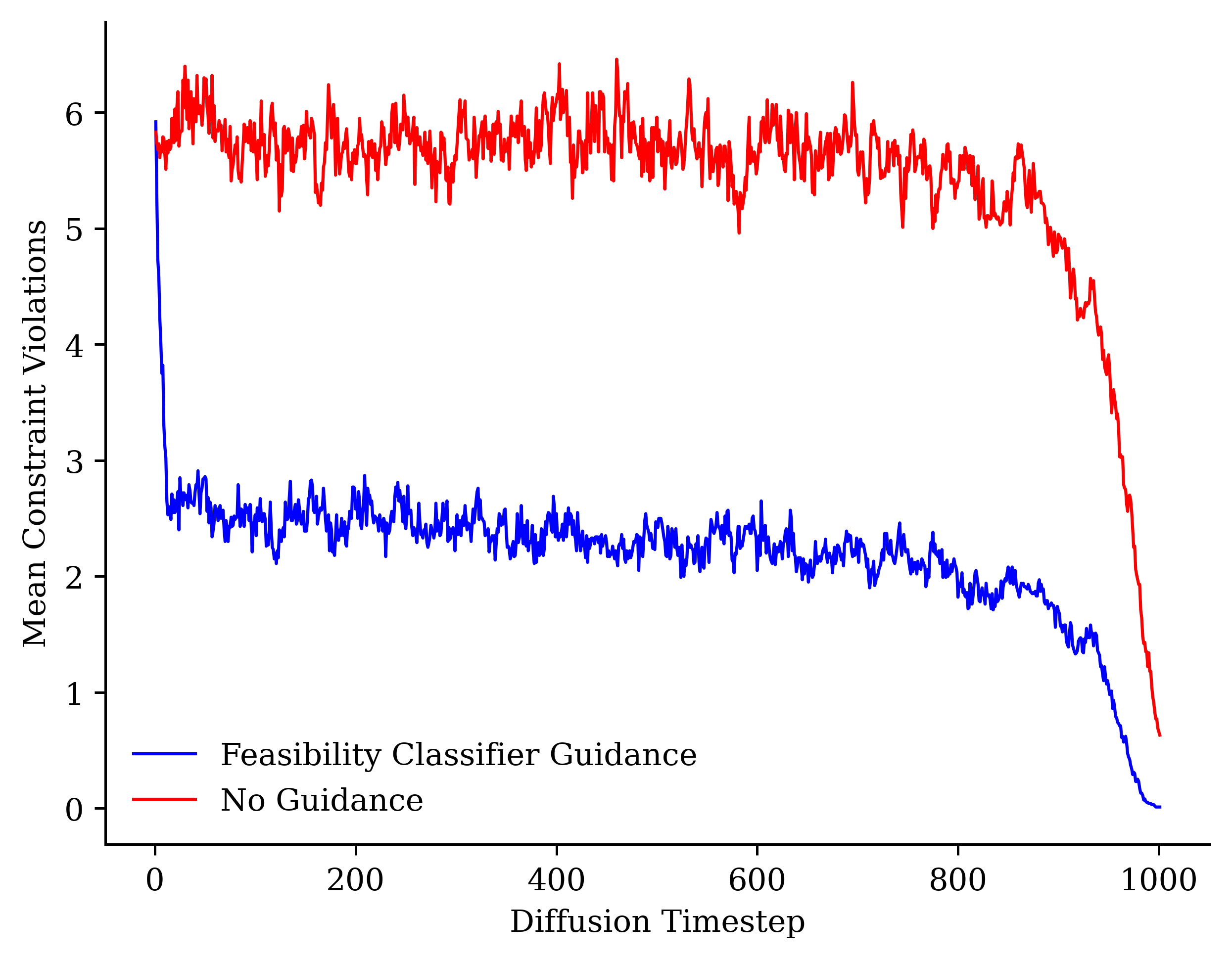}
\caption{Leveraging classifier guidance improves the feasibility of generated samples throughout the denoising process compared to the standard DDPM (no guidance). The classifier guidance is tuned to $\gamma = 0.5$.}
\label{fig:figure_HullFeasWithT} 
\end{center}
\end{figure}

While the standard DDPM generates feasible hulls with a relatively high success rate, feasibility can be improved by leveraging guidance from a pre-trained classifier neural network. The classifier identifies hulls as satisfying all the constraints or violating at least one of the forty nine constraints. This classifier network was implemented in the denoising step of generating samples with a standard DDPM to guide the generation of hulls towards satisfying the feasibility criteria. As mentioned in the Methods Section, the degree to which the guidance influences sample denoising is tied to a hyperparameter, $\gamma$. Figure~\ref{fig:figure_feasVsGamma} shows the percentage of generated feasible samples among generated samples versus $\gamma$. Note that when $\gamma = 0$, the denoising process is the same as the standard DDPM. In addition to design feasibility, $\gamma$ also affects the dataset coverage of the generated samples, as shown in Figure~\ref{fig:figure_CandRVsGamma}. As defined in the Methods Section, generated samples have increased realism with a generated sample by decreasing the Chamfer distance to its nearest neighbor belonging to the dataset of designs. Similarly, the generated samples have increasing dataset coverage with decreasing distance of every dataset point to its nearest neighbor belonging to the generated samples. Realism and coverage are measured as the mean normalized chamfer distance between the generated hulls and the dataset hulls. Table~\ref{table:gammaCoverage} quantifies coverage, showing that increasing $\gamma$ reduces the dataset coverage substantially. To maintain dataset coverage similar to the interpolation studies, $\gamma$ should be less than or equal to 0.35. To balance both design feasibility and dataset coverage among generated samples, $\gamma$ is set to 0.5. This way, feasible samples are generated 99.5\% of the time and maintain a dataset coverage ratio greater than 0.9. The remaining plots in this subsection capture a snapshot of samples generated with guided diffusion with $\gamma$ set to 0.2, 0.35, 0.5, 0.65, 0.80, and 1.0. Table ~\ref{table:gammaFease} shows the trend of both increasing success in generating feasible hulls. The PCA charts in Figure ~\ref{fig:figure_diff_guided_gamma_PCA} illustrate the reduction in coverage with increasing $\gamma$. Figure~\ref{fig:figure_HullFeasWithT} shows that classifier guidance has a significant influence on the feasibility of generated samples throughout the denoising process.

\begin{table}
\begin{center}
\begin{tabular}{l r r} 
\hline
\multirow{2}{*}{\textbf{Generation Method}} & \textbf{Chamfer Distance} & \textbf{Normalized Coverage} \\
 & \textit{(Lower is Better)} & \textit{(Higher is Better)} \\
\hline
Random Dataset Sample       & 4.315     & (Baseline) 1.000 \\
Interpolation Study 1       & 5.099     & 0.965 \\
Interpolation Study 2       & 2.976     & 1.059 \\
\\
\textbf{CTGAN}                       & \textbf{5.660}      & \textbf{0.940} \\
\textbf{Standard DDPM}  &  \textbf{4.672}   & \textbf{0.984} \\
\\
Guidance: $\gamma=0.2$      &  4.731    & 0.982 \\
Guidance: $\gamma=0.35$     &  5.067    & 0.967 \\
Guidance: $\gamma=0.5$	    &  6.002    & 0.925 \\
Guidance: $\gamma=0.65$     & 8.453     & 0.817 \\
Guidance :$\gamma=0.8$      & 13.611    & 0.589 \\
Guidance: $\gamma=1.0$      & 27.054    & -0.005 \\
\hline
\end{tabular}
\end{center}
\caption{The table provides the dataset coverage for the different sampling methods. These values are normalized between the best- and worst-case scenarios found in the dataset. The standard DDPM covers the dataset better than the CTGAN. By adding guidance to the DDPM, dataset coverage is maintained when $\gamma \leq 0.5$.}
\label{table:gammaCoverage}
\end{table}

\begin{table}
\begin{center}
\begin{tabular}{l r } 
\hline
%\multirow{2}{*}{\textbf{Generation Method}} & \multirow{2}{*}{\textbf{Feasible}}& \multicolumn{4}{c}{\textbf{Constraint Violations}}\\
% &  & \textbf{1} & \textbf{2} & \textbf{3}& \textbf{4+}\\
\textbf{Generation Method} & \textbf{Feasibility Rate} \\
\hline
Interpolation Study 1       & 0.931 \\ % & 0.047 & 0.019 & 0.003 & 0.001\\
Interpolation Study 2       & 0.938 \\ % & 0.037 & 0.022 & 0.002 & 0.000\\
 \\
\textbf{CTGAN}                       & \textbf{0.007} \\ % & 0.039 & 0.092 & 0.146 & 0.715 \\
\textbf{Standard DDPM}      & \textbf{0.511}  \\ %   & 0.276	& 0.124 & 0.063	& 0.026 \\
\\
Guidance: $\gamma=0.2$	    & 0.839	 \\ %    & 0.122	& 0.030  & 0.006	& 0.003 \\
Guidance: $\gamma=0.35$     & 0.962	 \\ %   & 0.030    & 0.007	  & 0.001	& 0.000 \\
Guidance: $\gamma=0.5$	    & 0.995  \\ % & 0.005     & 0.000   &	0.000   & 0.000 \\
Guidance: $\gamma=0.65$     & 1.000	 \\ %   & 0.000	    & 0.000	  & 0.000	& 0.000 \\
Guidance: $\gamma=0.8$	    & 1.000 \\ %	& 0.000	    & 0.000  &	0.000   & 0.000 \\
Guidance: $\gamma=1.0$	    & 1.000 \\ %	& 0.000	    & 0.000  &	0.000   & 0.000 \\
\hline
\end{tabular}
\end{center}
\caption{The table shows the fraction of generated samples that are feasible. The standard DDPM generates feasible samples 73x more often than CTGAN. Increasing $\gamma$ increases the proportion of feasible samples. All generated samples are feasible when $\gamma \geq 0.65$}
\label{table:gammaFease}
\end{table}

\subsection{Hull Generation with Performance Guided Denoising Diffusion Probabilistic Model}
In addition to generating feasible samples, guidance can also generate high-performing parametric hull designs. The following subsections provide the results from training performance prediction neural networks on seven objectives and the results from measuring and simulating hulls generated using multi-objective performance guidance. 

\subsubsection{Performance Prediction Training}
\begin{figure}[ht]
\begin{center}
\setlength{\unitlength}{0.012500in}%
\includegraphics[width=3.5in]{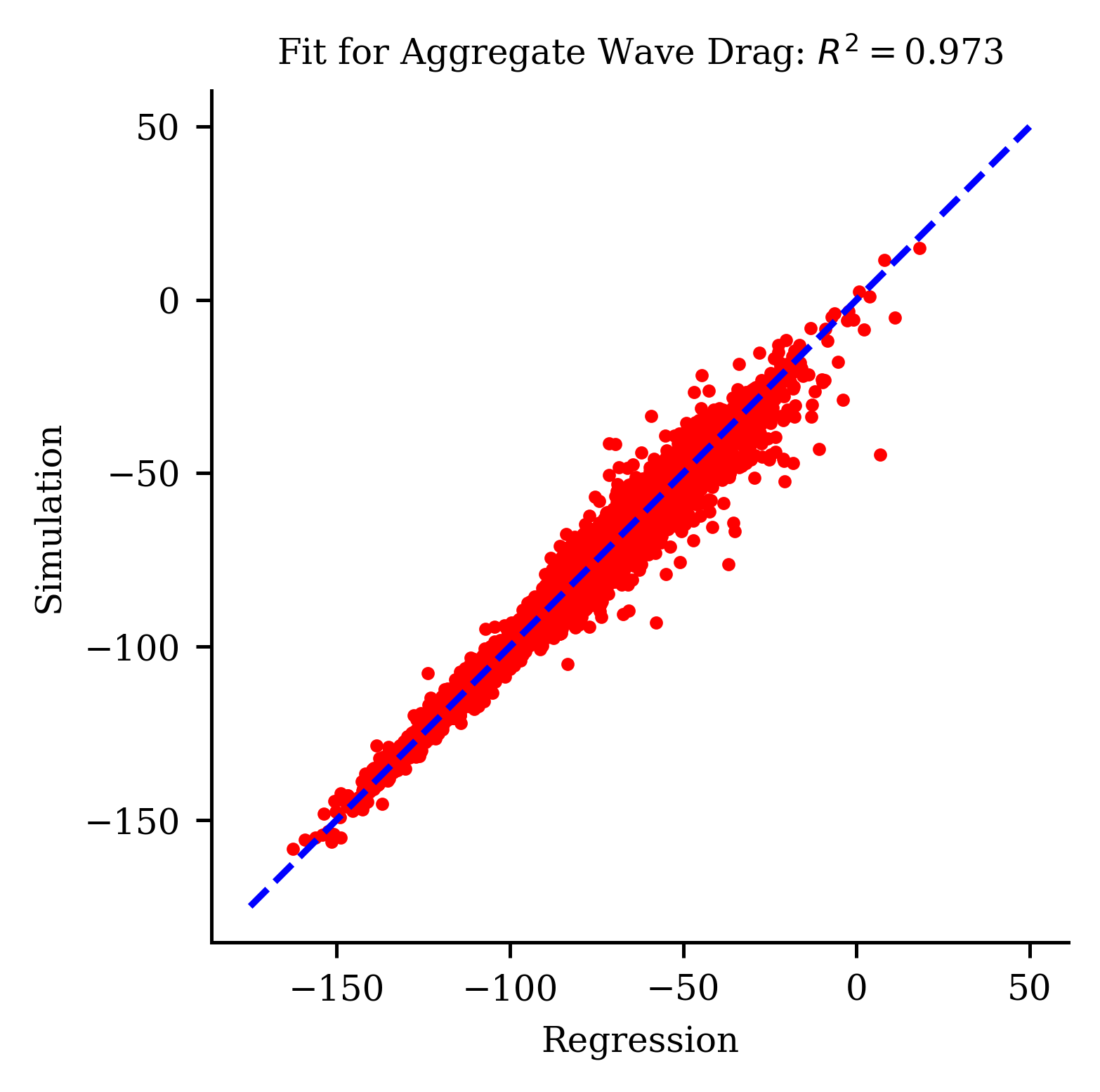}
\caption{Comparison of the neural network prediction to the simulation value (ground truth) across the dataset for aggregate wave drag. This regression had a $R^{2}$ equal to 0.973. A perfect prediction ($R^2 = 1$) is shown by the blue dashed line.}
\label{fig:figure_Cw_training} 
\end{center}
\end{figure}

\begin{figure}[ht]
\begin{center}
\setlength{\unitlength}{0.012500in}%
\includegraphics[width=5.5in]{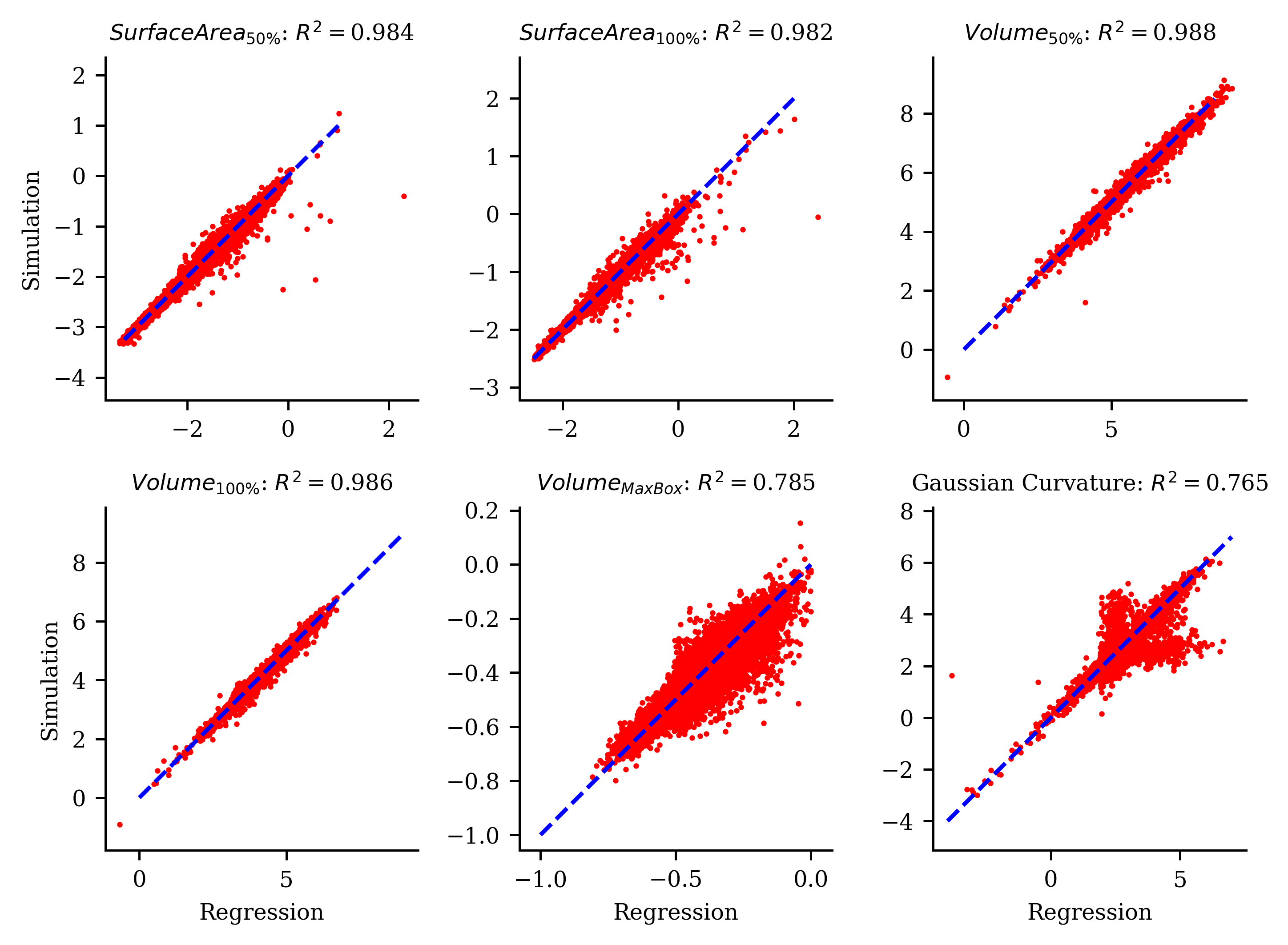}
\caption{Comparison of the neural network prediction to the simulation value (ground truth) across the dataset for the remaining six performance metrics. All of these performance metric regressions are well resolved. A perfect prediction ($R^2 = 1$) is shown by the blue dashed line.}
\label{fig:figure_OtherRegression_training} 
\end{center}
\end{figure}

Using the performance data from the Ship-D dataset, seven residual neural networks were trained to predict the performance of the hulls given the parameterized design vector. Table~\ref{table:PerfPredictionInfo} summarizes the results of the training, using $R^2$ as a measure of the goodness of fit for these neural networks. Figure~\ref{fig:figure_Cw_training} shows the plot of the regression prediction versus the simulation calculation for the aggregate wave drag measurement. Figure~\ref{fig:figure_OtherRegression_training} shows the same plots for the remaining six performance metrics. The blue dashed line in these figures represents the perfect regression prediction, exactly aligning with the simulation calculation. The wave drag coefficient, surface area, and volume prediction neural networks have high $R^2$ fits and hug the blue dashed line closely. The MaxBox and Gaussian Curvature predictions have lower $R^2$ values, however, they are still sufficient for use with performance-guided DDPM sampling~\cite{arechiga2023drag}.

\begin{table}[H]
\centering
\begin{tabular}{lc} 
\hline
\textbf{Performance Objective} & \textbf{Training Fit: [$R^2$]} \\
\hline
Wave Drag \(C_w\) & 0.973\\ 
Surface Area\(_{50\%}\) & 0.983 \\
Surface Area\(_{100\%}\) & 0.982 \\
Volume\(_{50\%}\)  & 0.988 \\
Volume\(_{100\%}\)  & 0.986 \\
Volume\(_{\text{MaxBox}}\) & 0.784 \\
Gaussian Curvature & 0.765\\
\hline
\end{tabular}
\caption{The performance prediction neural networks have high goodness-of-fits to the training data, which enables performance guidance in DDPM sampling with these objectives.}
\label{table:PerfPredictionInfo}
\end{table}

\subsubsection{Multi-objective Guided Performance Hull Generation}
\begin{figure}[ht]
\begin{center}
\setlength{\unitlength}{0.012500in}%
\includegraphics[width=3.5in]{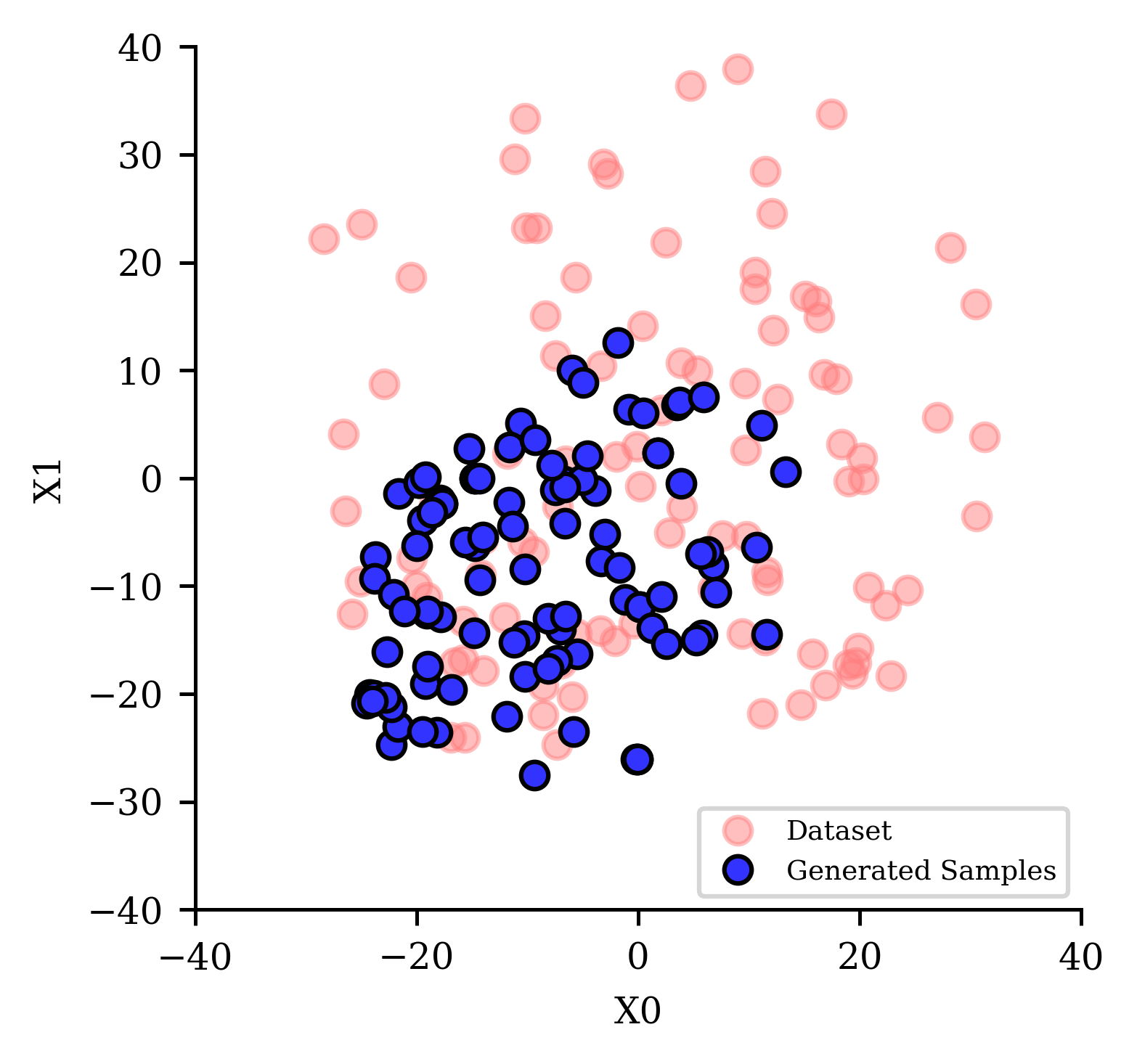}
\caption{Two-dimensional Principal Component Analysis of the hull parameterization shows that the performance-guided DDPM with $\gamma = 0.5$ leads to sample coverage that is skewed relative to the distribution of the Ship-D dataset as a result of the performance guidance.}
\label{fig:figure_PerfPCA} 
\end{center}
\end{figure}

The seven performance prediction neural networks were implemented with the guided DDPM to generate 1000 hulls. Each objective in these samples was randomly weighted so that the influence of each of the performance metrics varied among the samples. The feasibility classifier guidance was tuned to $\gamma$ = 0.5 to maintain some variability and dataset coverage among the samples and to not overpower the performance guidance. The samples generated with performance guidance were feasible 83.9\% of the time. The PCA plot of these generated samples is shown in Figure~\ref{fig:figure_PerfPCA}. These samples do not cover the sample range of the design space as the Ship-D dataset hulls. 

After sampling, the 839 feasible hull designs were simulated and measured with the seven performance objectives. The mean and standard deviation of the performance metrics among the Ship-D dataset and the generated samples are provided in Table~\ref{table:tabGeneratedSamples}. These metrics are scaled according to Equations~\ref{eq_NormCW}-~\ref{eq_NormGC}, so it is important to note that these values exist on a logarithmic scale. Among these samples, the wave drag coefficients and displaced volumes showed significant improvements in their performance. These improvements were at the expense of a relative increase in the surface area and Gaussian curvature. The generated samples have wave drag coefficients for any single speed/draft condition that is, on average, 91.4\% lower than the average wave drag coefficients of the Ship-D dataset hulls. For the displaced volumes, these generated hulls have an average 114x increase in displaced volume in the bottom 50\% of the hull depth and an average 47.9x increase in the total displaced volume of the hull. The generated hulls have, on average, 2.1x more total surface area, 4.4x more surface area in the bottom 50\% of the hull, and 1.51x more double curvature compared to the Ship-D hulls. This is not desirable. The MaxBox metric saw a small, but negligible decrease in the volume ratio of the hull belonging to the MaxBox, where the generated samples have an average 5.2\% reduction in the MaxBox volume ratio compared to the hulls in the dataset. This result, however, is far overshadowed by the substantial increase in total available volume in the hull.

\begin{figure}[ht]
\begin{center}
\setlength{\unitlength}{0.012500in}%
\includegraphics[width=5.5in]{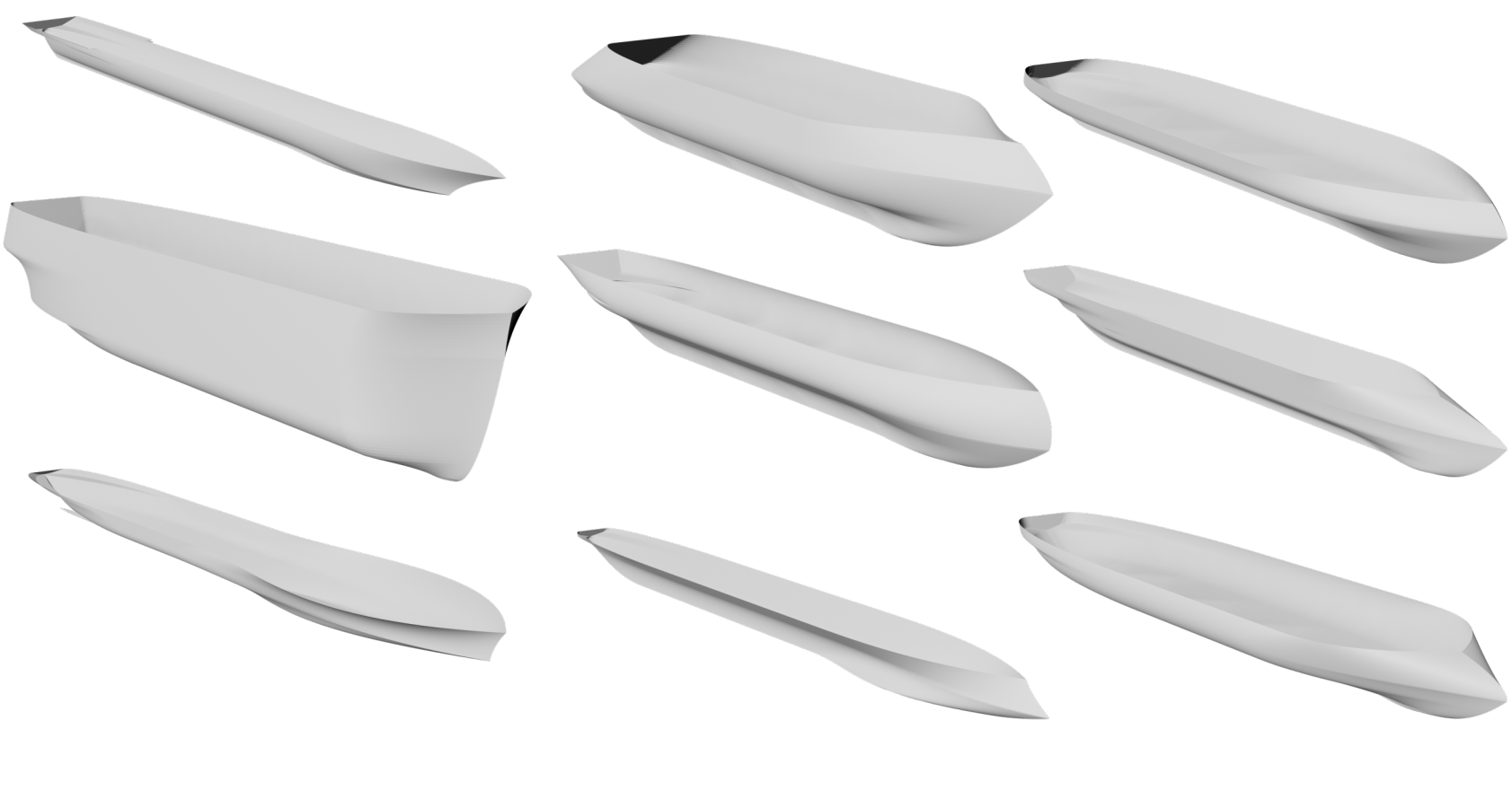}
\caption{A selection of hulls generated with multi-objective guided performance generation. Notice the relative slenderness of the hulls leading to drastically reduced drag coefficients relative to the dataset hulls.}
\label{fig:figure_GenHulls} 
\end{center}
\end{figure}

In addition to measuring the performance of these hulls, a .stl mesh and 5 images of each hull were created for visual analysis. Figure~\ref{fig:figure_GenHulls} shows nine of these hulls. A major difference in these generated hulls is their higher length-to-beam ratios compared to the Ship-D hulls seen in Figure~\ref{fig:figure_Ship-DHulls}.

\begin{table}
\begin{center}
\begin{tabular}{l r r r rr} 
\hline
\multirow{3}{*}{\textbf{Performance Objective}} & \multicolumn{2}{c}{\textbf{Ship-D Dataset}} & \multicolumn{2}{c}{\textbf{Generated Samples}} & \textbf{Scaled Factor} \\ 
& \multicolumn{2}{c}{($Y_{*gen}$)} & \multicolumn{2}{c}{($Y_{*DS}$)} & \multirow{2}{*}{($Y_{gen}/Y_{DS}$)}\\
 & Mean & Std. & Mean & Std. & \\
\hline
Wave Drag \(C_w\) & -73.40 & 17.38 & -107.45 & 23.90 & \textbf{0.086}\\ 
Surface Area\(_{50\%}\)  & -1.71 & 0.53 & -1.07 & 0.19 & 4.365\\
Surface Area\(_{100\%}\) & -1.09 & 0.45 & -0.76 & 0.19 & 2.138 \\
Volume\(_{50\%}\) & 4.78 & 0.81 & 2.72 & 0.59 & \textbf{114.815}\\
Volume\(_{100\%}\) & 3.80& 0.62 & 2.12 & 0.43 & \textbf{47.863}\\
Volume\(_{\text{MaxBox}}\)& -0.407 & 0.010 & -0.384 & 0.072 & 0.948\\
Gaussian Curvature& 2.43 &  0.529 & 2.61& 0.24 & 1.514\\
\hline
\end{tabular}
\end{center}
\caption{The table shows the mean and standard deviation of the performance metrics across the Ship-D dataset hulls and the feasible generated hulls. The generated hulls saw a 47.9x increase in total volume and a 91.4\% relative decrease in wave drag coefficient across all speeds.}
\label{table:tabGeneratedSamples}
\end{table}

%\begin{figure}[H]
%\begin{center}
%\setlength{\unitlength}{0.012500in}%
%\includesvg{images/Guided_Perf_ConstraintSatisfaction.svg}
%\caption{Bar graph showing the number of individuals generated with an increasing number of constraint violations. The performance-guided DDPM with $\gamma = 0.5$ leads to feasible samples being generated 83.9\% of the time.}
%\label{fig:figure_PerfFeas} 
%\end{center}
%\end{figure}

%%%%%%%%%%%%%%%%%%%%%%%%%%%%%%%%%%%%%%%%%%
\section{Discussion}
The following subsections provide insight into the results of the studies presented. The first subsection reviews the successful generation of feasible designs with the different DDPMs. The second subsection provides an analysis of the dataset coverage of the DDPMs, with special attention made to the $\gamma$ hyperparameter used in the classifier-guided DDPM. The third subsection analyzes the performance of the hulls generated with performance guidance. The Appendix contains a discussion on the conditional DDPM study. 

\subsection{Feasibility Constraint Satisfaction}
Of the different DDPMs, only the classifier-guided DDPM showed to successfully sample feasible parameterized hull designs with the same success rate as the interpolation study. The standard DDPM, while only producing a feasible hull approximately half of the time, was able to do so only by implicitly learning the statistical relationships between design parameters in feasible hulls. The standard DDPM's feasibility success rate of 51.1\% success rate is significantly higher than the success rate of 0.66\% seen by randomly sampling the design space. 
 In the comparison between CTGAN and DDPM, a significant performance gap was identified, particularly in constraint modeling. The CTGAN benchmark study revealed that the standard DDPM is two orders of magnitude more successful at generating feasible hulls than the CTGAN model. While the dataset coverage study demonstrated CTGAN's ability to produce parameterized vectors representing the dataset statistics, leading to high coverage, it struggled to generate these design vectors with combinations of parameter values that result in high feasibility. This suggests that CTGAN may face challenges in implicitly learning the statistical correlations between the parameters to the extent that DDPM does, underlying the need for a comprehensive examination. Such an examination, backed by empirical and theoretical analysis, is essential to delve deeper into the observed challenges and understand the inherent model characteristics or learning behaviors causing the performance disparity. Without this thorough analysis, making definitive claims regarding the observed differences remains speculative. A deeper comparative study on constraint satisfaction across different deep generative models is needed to make such claims.

 Finally, the guided classifier guidance showed that by tuning the $\gamma$ hyperparameter, the rate of feasible hull generation varied. In order to meet the feasibility benchmark of 93\% feasible hulls, the $\gamma$ should be set between 0.35 and 1. Table~\ref{table:gammaFease} also shows that simply by including a small influence of guidance ($\gamma=0.2$), the success of generating feasible hulls improves significantly compared to the standard DDPM. For performance guidance, the success rate of generating feasible hulls was 83.9\%, which is lower than the benchmark target, but this comes at the benefit of producing high-performing hulls, even with $\gamma = 0.5$. This reduction in the feasibility satisfaction rate is due to the added influence of the performance guidance, which does not consider feasibility when generating samples. Further work in hyperparameter tuning can lead to higher success rates in feasible and high-performing hull generation. Overall, a DDPM with classifier guidance can be used to generate feasible design vectors with a high degree of success in an extremely complicated design space. 

\subsection{Dataset Coverage}
Among the DDPMs, there were varying degrees of dataset coverage. Visually comparing the design space coverage of the Ship-D dataset hulls and the generated samples proved to be reasonably effective at analyzing the dataset coverage. Among the two interpolation methods, interpolation between a random design vector and its nearest neighbor was the best benchmark for dataset coverage as the diversity in the generated samples relied on the diversity of the randomly selected design vectors. The standard DDPM was also effective at maintaining dataset coverage as it was trained to generate sample hulls that are representative of the hulls in the dataset, as seen in Figure~\ref{fig:figure_diff_vanilla_PCA}. The standard DDPM also maintains dataset coverage better than CTGAN. Finally, the DDPM with classifier guidance heavily relied on $\gamma$ for maintaining dataset coverage. As $\gamma$ is increased to produce feasible hulls with a higher success rate, the dataset coverage of the feasible samples decreases. Figure~\ref{fig:figure_CandRVsGamma} was made to quantify the coverage and realism of the samples in addition to visually inspecting the PCA distribution of the generated samples. The best balance of maintaining a high feasibility success rate and dataset coverage was at $\gamma \approx 0.35 - 0.5$, as shown in Table~\ref{table:gammaCoverage}. With increasing $\gamma$ the generated samples lose diversity and cluster around the center of the PCA distribution. While the performance-guided DDPM was not intended to generate designs that cover the dataset, Figure~\ref{fig:figure_PerfPCA} suggests that these generated samples do maintain some diversity and dataset coverage. Overall, the classifier-guided DDPM is shown the maintain a large breadth of dataset coverage with careful tuning of its hyperparameters. 

\subsection{Performance Guidance}
The DDPM with performance guidance produced hulls with mixed results. The performance guidance created hulls with an average 91.4\% lower drag and 47.9x higher displaced volumes than the hulls from the original dataset. This is a highly desirable outcome of the performance guided sample generation. This outcome, however, came at the expense of the generated hulls having increased surface area and double curvature, which is not ultimately desirable. Further in-depth analysis, such as life cycle costs assessments, is needed to weigh the impact these results would have on a real, scaled-up ship instead of a non-dimensionalized parametric hull shape. Future work can also consider different tuning of the $\gamma$ and $\lambda$ hyperparameters in the performance guidance of the hulls. The random $\lambda$ weights used for this study were likely not scaled appropriately for the magnitude of the gradients of the different performance metrics. This could explain why the aggregate wave drag coefficient and displaced volume metrics were improved drastically; while MaxBox was relatively unaffected, and surface area and Gaussian curvature were increased. As these performance gradients were calculated using the weights of the regression neural networks,  it is possible that the magnitudes of gradients between the different regression models disproportionately affected the net influence of guidance on design generation. This is rather apparent when comparing the significant, yet desired, increase in hull volume and the undesired increase in surface area. These two competing performance objectives should have maintained some balance of improvement across the DDPM generated samples; however, the displaced volume performance objectives saw overwhelming improvement that included a relative detriment to the surface area objectives. Nonetheless, the significant improvements in wave drag coefficient and displaced volume have strong economic prospects on the cost of operating a ship: the cost of fuel (drag) and the ability to generate revenue (carry cargo). Leveraging a DDPM with performance guidance has been shown to generate hull shapes considering multiple objectives that can lead to huge cost savings to ship operators. Future work will explore generating hulls with specific performance requirements to find explicit applicability of guided DDPMs to generate hulls tied to real cost savings in ship design. 

In addition to the performance of the generated hulls, these generated samples share more semblance of real ship hulls than do the Ship-D hulls. Figure~\ref{fig:figure_GenHulls} shows a sample of the generated hulls for visual inspection. These hulls have higher length-to-beam ratios than the Ship-D hulls and have streamlines that are more akin to real ship hulls.

%%%%%%%%%%%%%%%%%%%%%%%%%%%%%%%%%%%%%%%%%%
\section{Conclusion}
The goal of this work was to generate ship hulls using a denoising diffusion probabilistic model that considers the performance of the hull as part of the design generation. First, by training a DDPM on a dataset of randomly generated feasible hull designs, the DDPM was able to implicitly learn statistical relationships between the design parameters to generate feasible parametric hulls. Then, by incorporating guidance from performance prediction models trained on the same dataset of hulls, the DDPM was able to generate high-performing hulls with only information learned from the low-performing hulls in the dataset. 

One critical aspect of leveraging generative AI on parametric design information is to generate feasible designs. A standard diffusion model can generate feasible hulls approximately 51.1\% of the time. While this is much better than the success rate seen by randomly generating hull parameterizations ($\sim0.66\%$), standard diffusion models do not yield feasible hulls at rates similar to interpolation methods ($\sim93\%$). By leveraging the gradients of a classifier during the sampling process, the standard DDPM saw increased feasibility among generated hull designs. The classifier guidance in the denoising process was influenced by a tunable parameter, $\gamma$. By varying $\gamma$, the guided DDPM was able to generate hulls with different success rates of feasibility at the expense of design coverage across the dataset. It was found that $\gamma$ = 0.5 led to high hull feasibility (99.5\%) with limited detriment to dataset coverage. 

As guidance was shown to be the most successful and versatile method of producing feasible hull designs, guidance was also used to generate high-performing designs. Seven neural networks were trained to predict the different performance metrics given a hull's design vector. The gradients of these performance prediction neural networks were implemented for performance guidance in the DDPM's sampling process. The aggregate wave drag coefficients of the generated hulls had a 91.4\% mean reduction in drag coefficient compared to the Ship-D hulls. The total displaced volume of the generated samples was on average 47.9x larger than the mean displaced volume of the Ship-D dataset hulls. However, surface area, Gaussian curvature, and MaxBox of the generated samples did not improve compared to the hulls in the dataset.  Overall, the significantly reduced drag coefficients and increased displaced volume are extremely beneficial to ship design.  These performance metrics dictate how expensive a voyage is (fuel costs due to drag) and how much cargo the ship can carry (how much money can be made on a voyage). With this work,  the economic prospect of leveraging generative AI to design ship hulls is shown.

\subsection{Future Work}
Future work will focus on the continued study of generative AI to generate ship hulls and other systems on a ship. Immediate future work will look at continued tuning of the $\lambda$ weights during performance guidance to generate hulls that have improved performance in all of the objectives. To accomplish this, a study on hyperparameter tuning and guidance gradients will be performed. Further, work in leveraging guided diffusion to generate high-performing ship hulls with specific performance targets will be explored. The goal of this future work is to generate hulls that consider specific user-defined constraints (such as dimensions, volume, speed, etc.) with high performance. This way, the design of ship hulls using DDPMs could be analogous to similarly structured online text-to-image DDPMs, such as \textit{Dall-E}~\cite{ramesh2022hierarchical} and \textit{Stable Diffusion}~\cite{rombach2021highresolution}. In addition to generating ship hulls, further work in DDPMs to generate other aspects of ship design will be explored, such as structural design generation, packing arrangements, machinery, and outfitting on a ship. 

%%%%%%%%%%%%%%%%%%%%%%%%%%%%%%%%%%%%%%%%%%

%%%%%%%%%%%%%%%%%%%%%%%%%%%%%%%%%%%%%%%%%%

\section{Acknowledgements}
This research is funded by the United States' Department of Defense, Office of Naval Research, via the National Defense Science and Engineering Graduate (NDSEG) Fellowship program. The authors would like to thank MIT Supercloud for providing some of the computational resources needed to perform this work~\cite{supercloud}. The data and code used for the ShipGen project are available at \url{https://decode.mit.edu/projects/ShipGen/}

\appendix

\section{APPENDIX: Conditional Diffusion Model}
The appendix contains an additional study performed by training a conditional DDPM. Conditional DDPMs are similar to the standard DDPM, however, their structure includes extra layers that embed extra information in the training and sampling process. The extra information in this study is a sample's classification of being feasible or invalid. The following subsections detail the Methods, Results, and a Discussion on leveraging a conditional DDPM to generate hull designs. 

\subsection[\appendixname~\thesubsection]{Methods}
A tabular DDPM was built with additional conditioning embedding layers to influence the model to produce designs that satisfy the feasibility constraints. In addition to the 30,000 parametric hulls in the Ship-D dataset, 20,000 design vectors were randomly generated that do not meet at least one of the feasibility constraints. The feasible ship hull design vectors and the infeasible vectors were labeled respectively. While training the tabular DDPM, the feasibility label was provided to the conditional embedding layer with its respective design vector in training. The goal of the conditioning is to use that additional label to influence the sampling process to guide the tabular DDPM to produce designs that satisfy the feasibility constraints. To modify the standard DDPM to become a conditional DDPM modify the gradient step (Step 5) in the DDPM training algorithm with Equation~\ref{eq_ConditionTrain}, where C is the conditional embedding layer that is concatenated to the first layer of the standard DDPM. In sampling, replacing Step 4 of the sampling algorithm with Equation~\ref{eq_ConditionSample}, where C is the same conditional embedding layer concatenated to the first layer of the DDPM. The DDPM training and sampling algorithms are provided in Table~\ref{table:DDPMTrainAlg} and Table~\ref{table:DDPMSampAlg}, respectively.  The Results Section provides the sample distribution and constraint satisfaction of ship hull design vectors generated with the conditioned tabular DDPM. 

\begin{equation}
\nabla_{\theta}||\epsilon - \epsilon_{\theta}(\sqrt{\bar{a_t}}X_0 + \sqrt{1 - \bar{a_t}}\epsilon, t, C)||^2
\label{eq_ConditionTrain}
\end{equation}

\begin{equation}
 X_{t-1} = \frac{1}{\sqrt{\alpha_t}} (X_t - \frac{1 - \alpha_t}{\sqrt{1 - \bar{\alpha_t}}} \epsilon_{\theta}(X_t ,t,C)) + \sigma_tZ 
\label{eq_ConditionSample}
\end{equation}

\subsection[\appendixname~\thesubsection]{Results}
\begin{figure}[H]
\begin{center}
\setlength{\unitlength}{0.012500in}%
\includegraphics[width=3.5in]{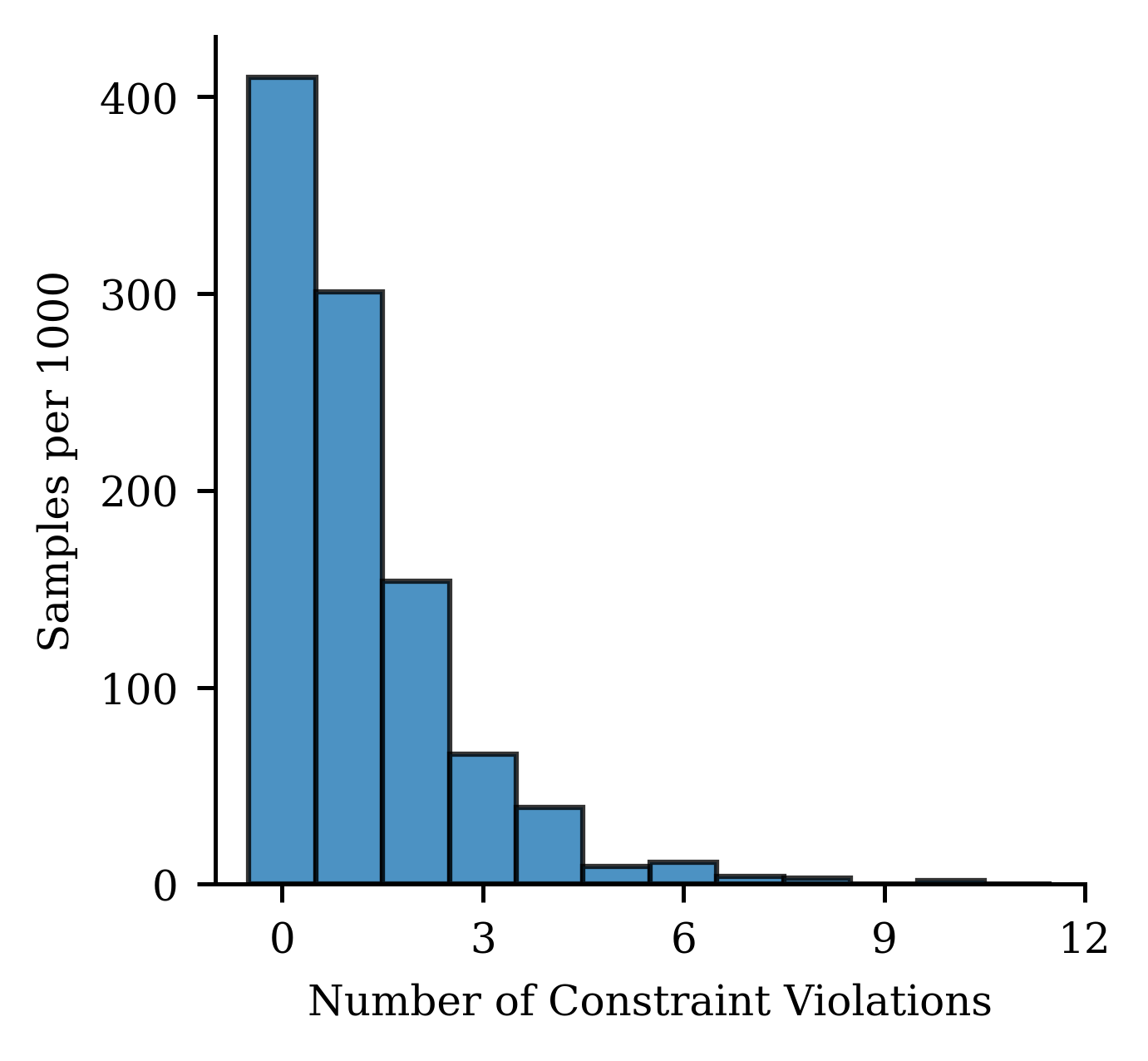}
\caption{Bar graph showing the number of individuals generated with an increasing number of constraint violations. Leveraging a conditional DDPM to generate hull parameterization leads to hull parameterizations that satisfy all constraints 39.8\% of the time}
\label{fig:figure_diff_conditional_bar} 
\end{center}
\end{figure}

\begin{figure}[H]
\begin{center}
\setlength{\unitlength}{0.012500in}%
\includegraphics[width=3.5in]{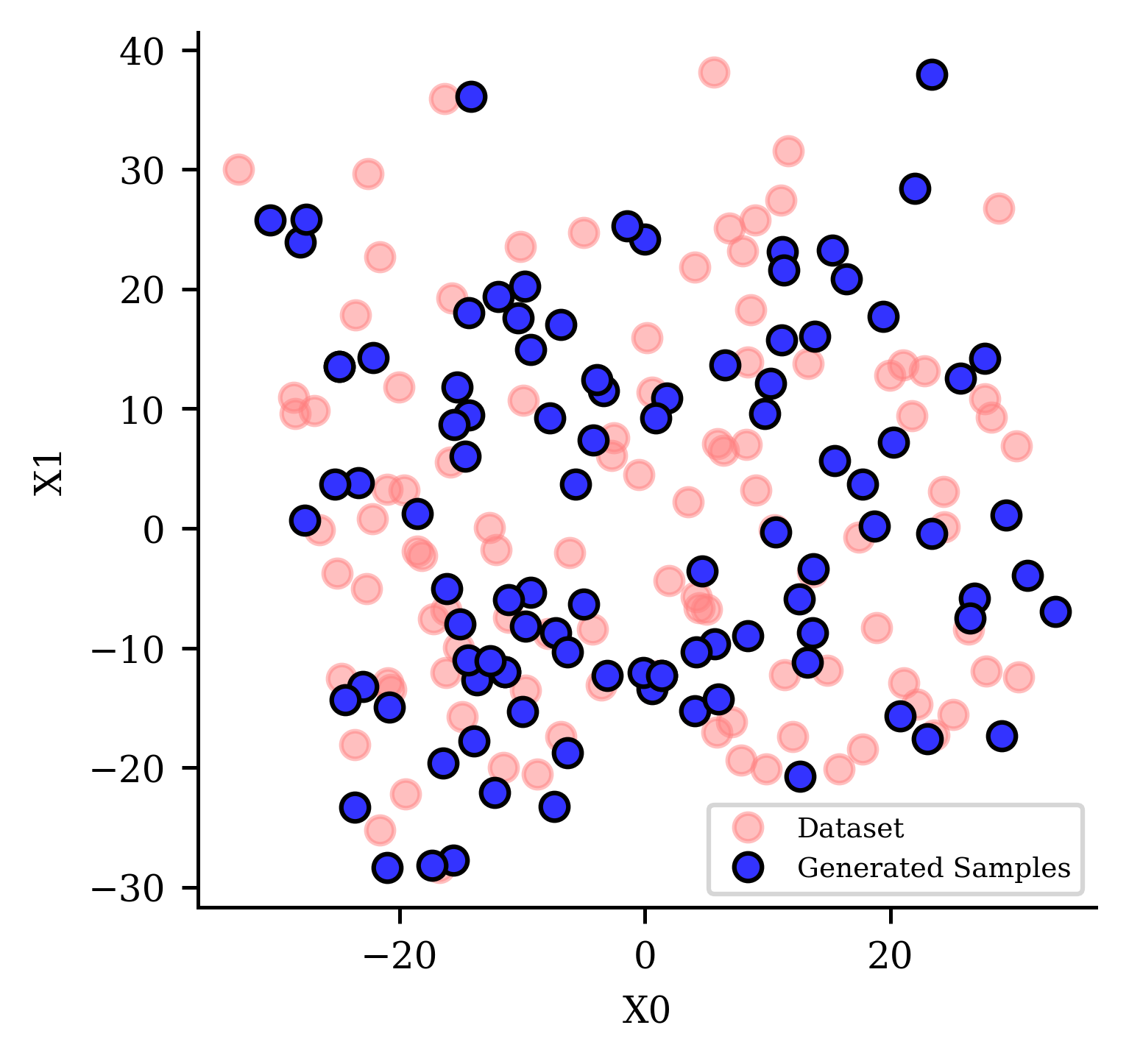}
\caption{Two-dimensional Principal Component Analysis of the hull parameterization shows that hulls generated with a conditional DDPM maintain most of the dataset coverage; however, there is less design feasibility among the samples than desired.}
\label{fig:figure_diff_conditional_PCA} 
\end{center}
\end{figure}

\begin{figure}[H]
\begin{center}
\setlength{\unitlength}{0.012500in}%
\includegraphics[width=3.5in]{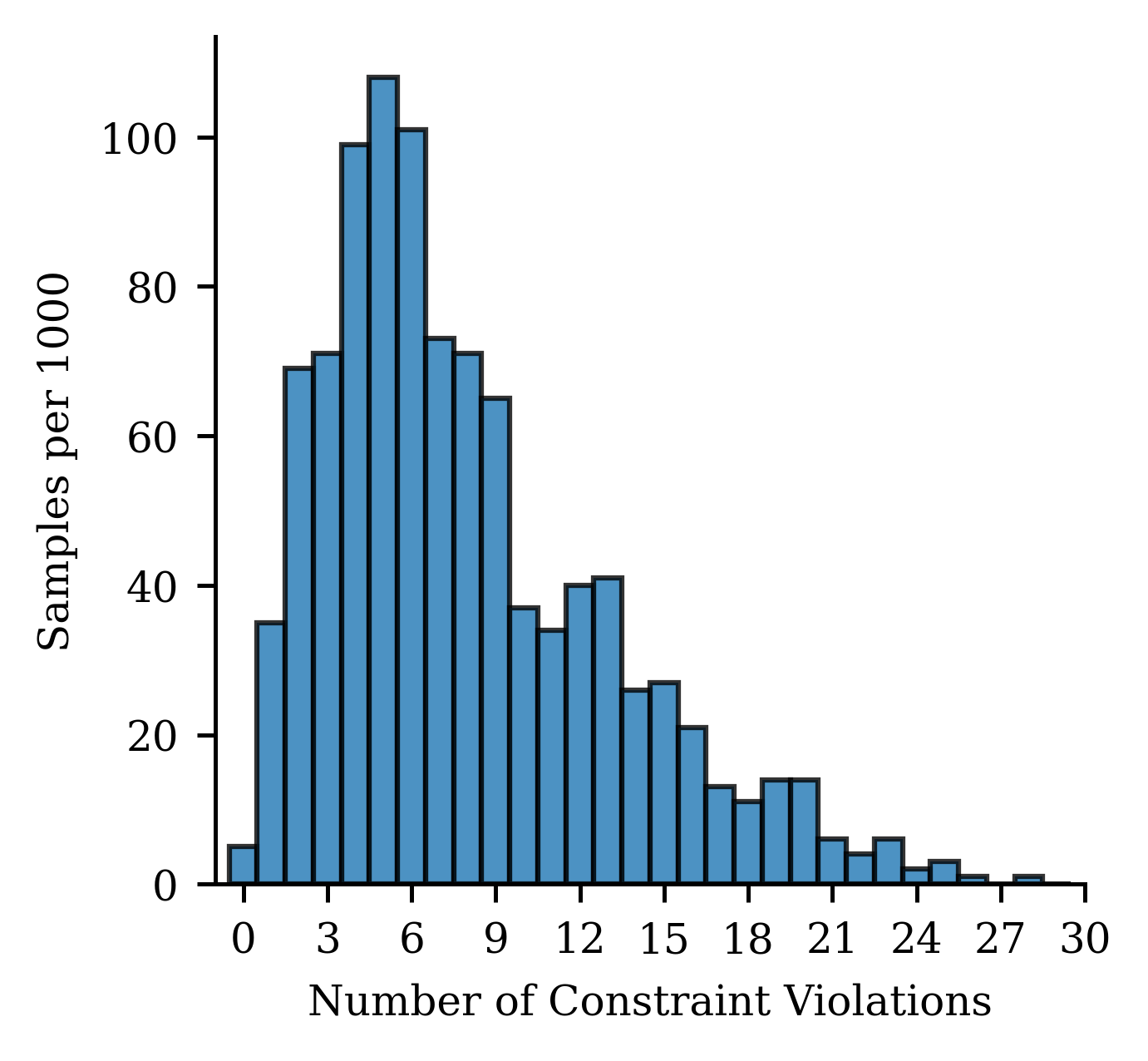}
\caption{Bar graph showing the number of individuals generated with an increasing number of constraint violations. Leveraging a conditional DDPM to generate invalid hull parameterization leads to the generation of samples that violate a large spread of a number of design feasibility constraints.}
\label{fig:figure_diff_negconditional_bar} 
\end{center}
\end{figure}

\begin{figure}[H]
\begin{center}
\setlength{\unitlength}{0.012500in}%
\includegraphics[width=3.5in]{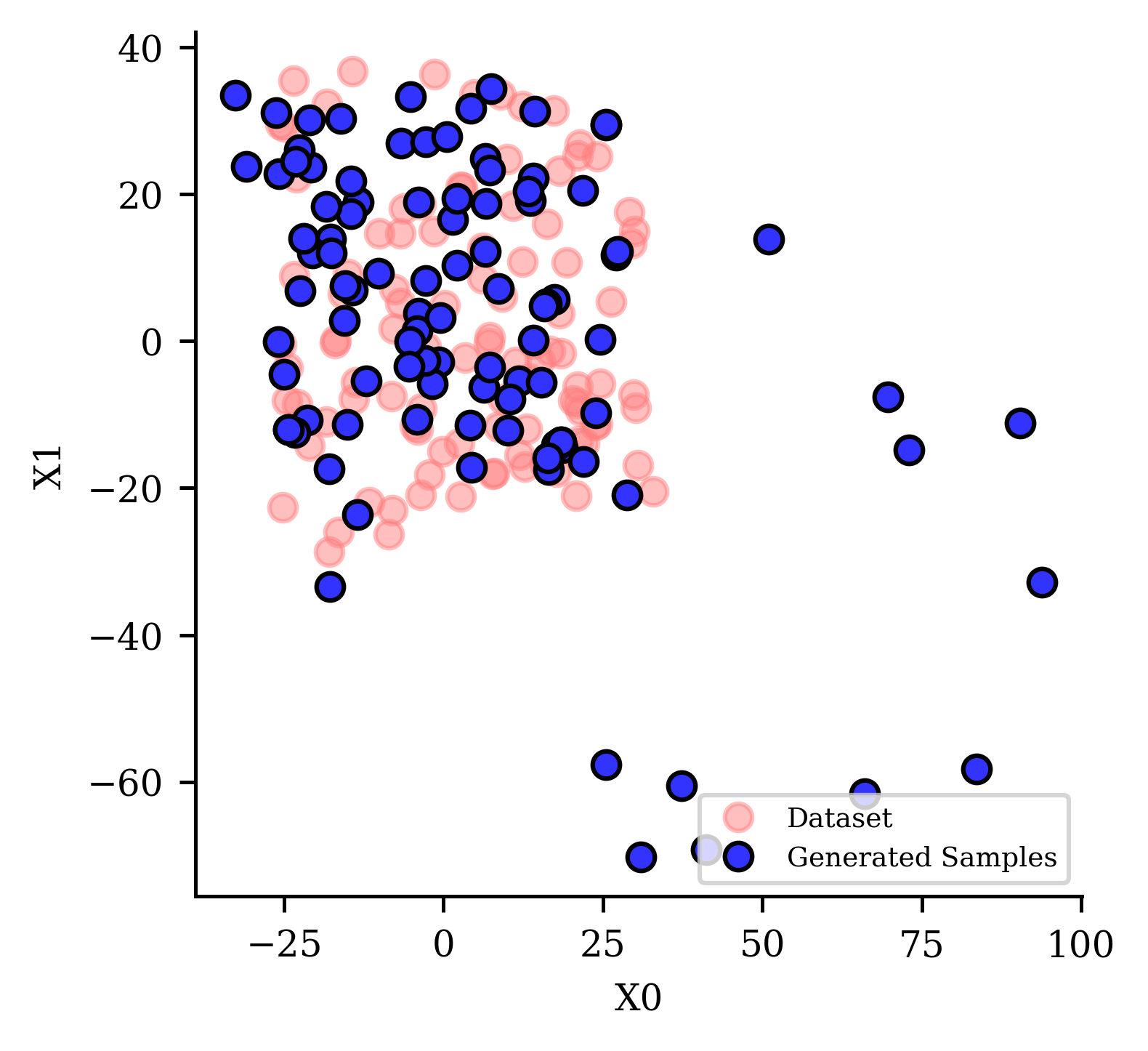}
\caption{Two-dimensional Principal Component Analysis of the hull parameterization shows that invalid hull samples cover the dataset space and more, as some of the parameters are intentionally sampled outside of the feasible range of some parameters.}
\label{fig:figure_diff_Negconditional_PCA} 
\end{center}
\end{figure}

The conditional DDPM was trained on the thirty thousand Ship-D hulls that satisfy all forty nine constraints and twenty thousand invalid samples that violate varying numbers of the forty nine constraints. Two separate sample generations were performed with the conditional DDPM. The first study tried to intentionally generate feasible hulls. Figure~\ref{fig:figure_diff_conditional_bar} shows that the conditional DDPM produces feasible hulls 39.8\% of the time. The PCA plot in Figure~\ref{fig:figure_diff_conditional_PCA} shows that the spread of the generated samples is within the bounds of the dataset.  This conditional DDPM can also intentionally generate invalid samples. Although this is not useful in design work, the generation of invalid samples with the conditional DDPM shows that the model can distinguish between ``positive'' and ``invalid'' samples in sample generation. Figures~\ref{fig:figure_diff_negconditional_bar} and ~\ref{fig:figure_diff_Negconditional_PCA} showcase the results of hulls that were intentionally created to violate the hull parameterization's feasibility criteria. 

\subsection[\appendixname~\thesubsection]{Discussion}
The conditional DDPM was not able to produce feasible hulls at a rate that meets the interpolation study benchmarks. It was surprising to see how poorly the conditional DDPM performed given that the model was trained with distinction between feasible and infeasible design vectors. The conditional model, however, could intentionally create infeasible hulls at will, which further adds to the surprise of the poor feasible hull generation. It seems that the conditional DDPM struggled to distinguish the statistical relationships between the parameters in the feasible and infeasible hulls during training. 
This result did not yield improvement over the standard DDPM in generating feasible hulls, but it was excellent in generating a large diversity of hulls with at least one constraint violation. This could be useful in future studies concerning design feasibility.

\section{APPENDIX: Parametric Hull Design: Parameters and Constraints}
\begin{figure}[H]
\begin{center}
\setlength{\unitlength}{0.012500in}%
\includegraphics[width=5.5in, page=1]{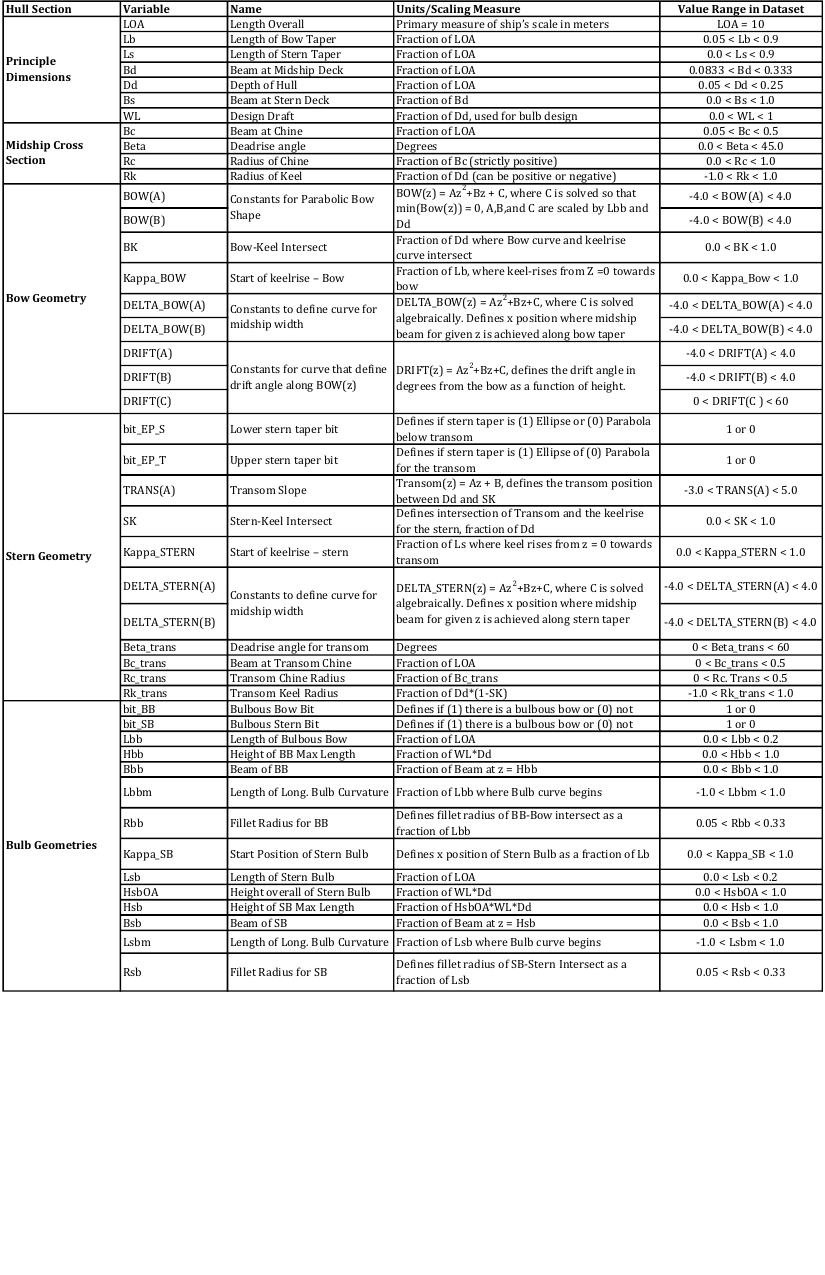}
\caption{List of the hull design parameters, their scaling, and their value ranges within the dataset.}
\label{fig:figure_PARAMETERS} 
\end{center}
\end{figure}

\begin{figure}[H]
\begin{center}
\setlength{\unitlength}{0.012500in}%
\includegraphics[width=5.5in, page=1]{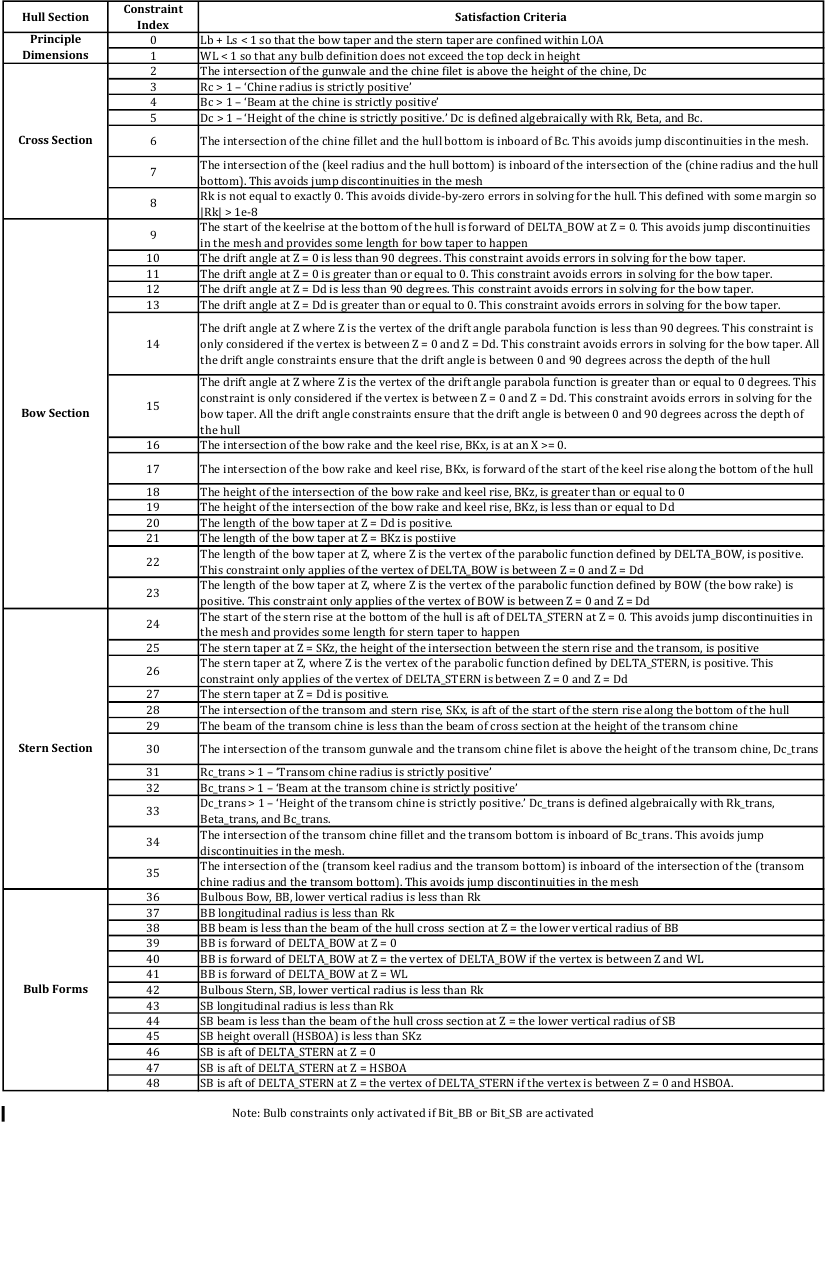}
\caption{List of the parametric hull design constraints and a description of their satisfaction criteria}
\label{fig:figure_CONSTRAINTS} 
\end{center}
\end{figure}

This appendix provides documentation for the 45 hull design parameters and the 49 algebraic feasibility constraints. Figure~\ref{fig:figure_PARAMETERS} lists the design parameters, describes the features, provides the ranges for each parameter in the Ship-D Dataset. Figure~\ref{fig:figure_CONSTRAINTS} lists each of the 49 algebraic constraints, and describes the conditions to satisfy each constraint. By satisfying all 49 algebraic constraints, the hull will satisfy the two feasibility criteria:

\begin{enumerate}
    \item The hull is watertight, meaning that there are no holes on its surface.
    \item The hull surface is not self-intersecting.     
\end{enumerate}

\printbibliography
%\bibliography{paper}  %%% Uncomment this line and comment out the ``thebibliography'' section below to use the external .bib file (using bibtex) .





% \end{thebibliography}

\end{document}